\pdfoutput=1
\def\Anonymised{false}
\def\MergeSupplement{true}
\documentclass[pdflatex,sn-vancouver-num]{sn-jnl}

\providecommand{\Anonymised}{false}
\providecommand{\MergeSupplement}{false}
\newif\ifanon            \csname anon\Anonymised\endcsname
\newif\ifmergesupp       \csname mergesupp\MergeSupplement\endcsname

\hypersetup{bookmarksdepth=2}
\AtBeginDocument{}
\usepackage{graphicx}
\usepackage{multirow}
\usepackage{amsmath,amssymb,amsfonts}
\usepackage{amsthm}
\usepackage{mathrsfs}
\usepackage[title]{appendix}
\usepackage{xcolor}
\usepackage{textcomp}
\usepackage{booktabs}
\usepackage{longtable}
\newsavebox{\fitwidthbox}
\newenvironment{fitwidth}{\begin{lrbox}{\fitwidthbox}}{\end{lrbox}%
  \ifdim\wd\fitwidthbox>\linewidth\resizebox{\linewidth}{!}{\usebox{\fitwidthbox}}\else\usebox{\fitwidthbox}\fi}
\newenvironment{widetable}[1][htbp]{\ifcsname tableorg\endcsname\tableorg[#1]\centering
  \else\table[#1]\centering\fi}{\ifcsname endtableorg\endcsname\endtableorg\else\endtable\fi}
\usepackage{refcount}
\RequirePackage[abbreviations,shortcuts=true]{glossaries-extra}
\setabbreviationstyle{long-short}

\makeatletter
\NewDocumentCommand{\smref}{mm}{%
  \ifmergesupp
    \ref{#2}%
    \begingroup
      \def\sm@kept{#1}\edef\sm@actual{\getrefnumber{#2}}%
      \IfRefUndefinedExpandable{#2}{}{\ifx\sm@kept\sm@actual\else
        \PackageWarning{smref}{Supplement pointer `#2' is stale: hand-kept
          `#1' but the merged build numbers it `\sm@actual'}%
      \fi}
    \endgroup
  \else
    #1%
  \fi}
\makeatother
\newcommand{\smNVAthree}{\smref{A}{app:nva_3}} %
\newcommand{\smNVAthirty}{\smref{B}{app:nva_30}} %
\newcommand{\smFullThirteen}{\smref{C}{app:full_13}} %
\newcommand{\smFullOneTwenty}{\smref{D}{app:full_120}} %
\newcommand{\smMethods}{\smref{E}{app:methods}} %
\newcommand{\smBlock}{\smref{E.1}{app:methods_block}} %
\newcommand{\smLambda}{\smref{E.2}{app:methods_lambda}} %
\newcommand{\smBaseClust}{\smref{E.3}{app:methods_baseclust}} %
\newcommand{\smMetrics}{\smref{E.4}{app:methods_metrics}} %
\newcommand{\smTransitions}{\smref{E.5}{app:methods_transitions}} %
\newcommand{\smRobust}{\smref{F}{app:unk}} %
\newcommand{\smScale}{\smref{F.5}{app:feature_scale}} %
\newcommand{\smBaseline}{\smref{G}{app:baselines}} %
\newcommand{\mainFigArchi}{\smref{2}{fig:archi}}
\newcommand{\mainTabHyper}{\smref{1}{tab:sr_hyperparams}}
\newcommand{\mainFigAriNVA}{\smref{7}{fig:ari_nmi_nva}}
\newcommand{\mainFigAriFull}{\smref{10}{fig:ari_nmi_full}}
\newcommand{\zenodoDOI}{10.5281/zenodo.22256000}
\newcommand{\codeURL}{https://github.com/MathisImm/sr-wordclasses}
\newcommand{\codeDOI}{10.5281/zenodo.23691971}  %
\newcommand{\reviewCodeURL}{TODO-anonymous-code-URL} %
\newcommand{\reviewDataURL}{TODO-anonymous-data-URL} %
\newabbreviation{rl}{RL}{reinforcement learning}
\newabbreviation{sr}{SR}{successor representation}
\newabbreviation{td}{TD}{temporal difference}
\newabbreviation[plural=PCAs]{pca}{PCA}{principal component analysis}
\newabbreviation{ppmi}{PPMI}{positive pointwise mutual information}
\newabbreviation[plural=UMAPs]{umap}{UMAP}{uniform manifold approximation and projection}
\newabbreviation{pos}{POS}{part-of-speech}
\newabbreviation[plural=MSEs]{mse}{MSE}{mean squared error}
\newabbreviation[plural=KLs]{kl}{KL}{Kullback-Leibler}
\newabbreviation{bgm}{BGM}{Bayesian Gaussian mixture model}
\newabbreviation{ari}{ARI}{Adjusted Rand Index}
\newabbreviation{nmi}{NMI}{Normalised Mutual Information}
\glsdisablehyper

\begin{document}

\title[Word-Class and Construction-Like Structure in Neural Successor Representations]{Word-Class and Construction-Like Structure Emerges in Neural Successor Representations Trained on Natural Language}

\ifanon\else

\author*[1]{\fnm{Mathis} \sur{Immertreu}}\email{mathis.immertreu@fau.de}

\author[2,3,4]{\fnm{Achim} \sur{Schilling}}

\author[2,4]{\fnm{Thomas} \sur{Kinfe}}

\author[1,2,3,4]{\fnm{Patrick} \sur{Krauss}}

\affil[1]{\orgdiv{Cognitive Computational Neuroscience Group, Pattern Recognition Lab}, \orgname{Friedrich-Alexander-University Erlangen-Nürnberg (FAU)}, \orgaddress{\street{Martensstr. 3}, \postcode{91058} \city{Erlangen}, \country{Germany}}}
\affil[2]{\orgdiv{Neuromodulation and Neuroprosthetics}, \orgname{University Hospital Mannheim, University Heidelberg}, \orgaddress{\street{Theodor-Kutzer-Ufer 1-3}, \postcode{68167} \city{Mannheim}, \country{Germany}}}
\affil[3]{\orgdiv{Neuroscience Lab}, \orgname{University Hospital Erlangen}, \orgaddress{\street{Waldstraße 1}, \postcode{91054} \city{Erlangen}, \country{Germany}}}
\affil[4]{\orgname{BG Clinic Ludwigshafen}, \orgaddress{\street{Ludwig Guttmann-Straße 13}, \postcode{67071} \city{Ludwigshafen}, \country{Germany}}}

\fi

\abstract{Neural language models are typically trained on next-token prediction, although linguistic structure spans multiple temporal scales. Successor representations (SRs) make this horizon explicit by encoding discounted distributions over future states. Here, we ask whether such predictive representations can recover not only word classes, but also finer functional and construction-like structure from natural language. A residual network trained on WikiText-103 predicts SR distributions at three horizons without part-of-speech supervision. At the shortest horizon, unsupervised clustering robustly recovers nouns, verbs, and adjectives, while directed inter-cluster transitions reproduce familiar syntactic asymmetries. At finer resolutions and across 13 part-of-speech categories, the same geometry reveals semantic-functional groupings that cross category boundaries and directed relations tracing candidate date, measurement, and title–name constructions. Part-of-speech agreement declines as the predictive horizon lengthens. These results suggest that word classes are coarse regions within a richer predictive geometry in which categorical and construction-like linguistic structure emerge from future-word distributions.}


\maketitle

\section{Introduction}\label{sec1}

Prediction provides a productive point of contact between artificial language models and theories of biological cognition. Neural language models are commonly trained to predict the next token, an objective that has supported representation learning from early neural probabilistic models to contemporary transformers~\citep{bengio2003neural,vaswani2017attention}. Although the training target concerns the next token, contextual architectures can use long preceding contexts and develop internal representations related to syntax and semantics. In neuroscience, prediction has likewise been proposed as a general organising principle of cortical processing~\citep{friston2010free,clark2013whatever}. A prominent complementary proposal treats the hippocampal formation as a predictive map that represents not only the next state, but the distribution of states expected over the future~\citep{stachenfeld2017hippocampus}, a proposal that ties prediction to the structure most closely associated with the formation and organisation of memory~\citep{eichenbaum2000cortical}.

This predictive perspective bears on a long-standing question in language acquisition: which aspects of grammatical organisation require language-specific prior constraints, and which can emerge from distributional experience? Nativist accounts have emphasised an innate language faculty and constraints on the hypothesis space available to learners~\citep{chomsky1965aspects}, whereas usage-based and constructionist accounts explain grammatical categories as abstractions over recurrent patterns of language use~\citep{tomasello2003constructing,bybee2010language}, with constructions themselves characterised as clusters of partially overlapping, lossy memory traces aligned by shared form, function, and context~\citep{goldberg2019explain}. The present study does not attempt to adjudicate the full debate between these traditions. It addresses a narrower computational question: can a predictive learner form word-class-like and construction-like representations without category supervision?

The idea that linguistic categories can emerge from distributional context has a longer connectionist history. Ritter and Kohonen showed that self-organising maps can arrange words topographically from the statistics of the contexts in which they occur, so that familiar linguistic categories emerge as neighbouring regions without category labels being supplied \citep{ritter1989self}. Elman subsequently demonstrated that a recurrent network trained to predict the next word in simple artificial sentences develops internal representations separating nouns from verbs and further subdividing them according to distributional and semantic properties; later work extended this predictive approach to grammatical relations and constituent structure \citep{elman1990finding,elman1991distributed}. Despite their different architectures, these studies share a common principle: lexical categories need not be imposed as discrete labels, but can arise geometrically from the relational structure of linguistic experience.

Previous work provides evidence that local linguistic prediction can induce such organisation. \citet{surendra2023word} trained recurrent neural networks to predict the next word from nine-word contexts in a natural-language corpus and found that hidden representations became organised by the \gls{pos} category of the predicted word, despite the absence of a \gls{pos} training objective. A subsequent extension replicated this pattern across German and English and found structure related to two-word \gls{pos} combinations~\citep{krauss2024word}. These results establish that category information can emerge from next-word prediction and its extension to two-word combinations. They do not, however, determine whether an explicitly multi-step predictive objective organises individual word types according to the longer-run transition structure of language.

\Glspl{sr} offer such an objective. Introduced in reinforcement learning, an \gls{sr} describes the expected discounted occupancy of future states rather than only the immediate successor~\citep{dayan1993improving,gershman2018successor}. Applied to language, a word is represented by the entire discounted distribution of words that follow it. The contrast with next-word prediction therefore lies in the target rather than the input: a next-token model may condition on a long preceding span, but it is still scored against a single following word, so longer-range structure is a by-product of context and depth. Here the discount factor sets that span explicitly, making the predictive horizon an experimental variable rather than a fixed property of the architecture, and training several horizons at once yields the same vocabulary at three settings of that variable.

\ifanon Earlier studies \else Our earlier studies \fi established several components of this approach in restricted domains. Neural \gls{sr} models learned spatial maps and the predefined transition structure of a small artificial language, in which word-class clusters emerged from 40 word states and three constructions~\citep{stoewer2022neural}. Related work used \glspl{sr} to form multiscale maps of handcrafted semantic features, semantic word embeddings, and multimodal image--word associations~\citep{stoewer2023neural,stoewer2023conceptual,stoewer2025multi}. Together, these studies showed that neural \glspl{sr} can recover spatial, semantic, linguistic, and multimodal structure. They left open the central question addressed here: whether \glspl{sr} trained directly on a large natural-language corpus recover grammatical and construction-like organisation from corpus-derived transition dynamics.

We therefore train a deep residual neural network on WikiText-103 to predict \gls{sr} distributions for a 20{,}000-token vocabulary at three temporal horizons. We analyse the resulting type-level representations without using \gls{pos} labels during model optimisation or clustering; \gls{pos} annotations are introduced only for post-hoc evaluation. The analysis proceeds from broad word classes to fine-grained lexical groupings and directed relations between clusters.

The central result is that multi-horizon successor prediction yields a structured linguistic geometry. At short horizons, nouns, verbs, and adjectives are recovered by unsupervised clustering, and the directed mass between their clusters shows descriptive asymmetries compatible with familiar syntactic relations, most clearly in the adjective's orientation towards nouns rather than verbs. At finer resolutions, the geometry separates lexical and semantic-functional subclasses and provides exploratory construction-like candidates for dates, measurements, and title--name sequences. \Gls{pos} agreement decreases at longer predictive horizons, showing that local category distinctions are most prominent when nearby successors are weighted strongly.

The contribution is thus a synthesis of two previously separate lines of work: word-class structure emerging from next-word prediction, and predictive cognitive maps learned through \glspl{sr}. To our knowledge, this is the first large-scale study to train neural multi-horizon \glspl{sr} directly on a naturalistic language corpus and evaluate their word-class, lexical, and construction-like organisation. The findings provide a proof of principle for the emergence of grammatical abstractions from predictive sequence learning.

\section{Methods}\label{sec2}

The methods fall into two parts. We first describe the \gls{sr}, the temporal-difference targets used to learn it, the network architecture, and the training setup; we then describe the post-hoc analysis pipeline, comprising the evaluation vocabulary and \gls{pos} lexicon, the consensus clustering procedure, the clustering metrics, and the inter-cluster transition networks. The full architecture specification, the composition of the training target, the base clustering settings, the formal metric definitions, and the transition-weight derivation are given in Supplementary Appendix~\smMethods{}.

\subsection{Successor Representation}

The \gls{sr}, introduced by \citet{dayan1993improving}, captures how future states are expected to unfold under a fixed policy, independently of the particular reward function. For a Markov decision process over states $s \in \mathcal{S}$ under a fixed policy $\pi$, the \gls{sr} $M(s,s')$ is defined as the expected discounted future occupancy of state $s'$ when starting from $s$:
\begin{equation}
M(s, s') = \mathbb{E}_\pi \left[\sum_{t=0}^{\infty} \gamma^t \mathbb{I}(s_{t+1} = s') \,\middle|\, s_0 = s\right],
\end{equation}
where $\gamma \in [0,1]$ is a discount factor and $\mathbb{I}(\cdot)$ denotes the indicator function. Intuitively, $M(s,s')$ quantifies how often, in a discounted sense, state $s'$ is expected to be reached when starting from $s$, with $\gamma$ setting the time horizon over which future states contribute. We adopt the arrival \gls{sr} formulation, predicting occupancy from the next state onwards, as the current state is already observed and carries no predictive value.

\subsection{Successor Representation Updates}

\begin{figure}
    \centering
    \includegraphics[width=\linewidth]{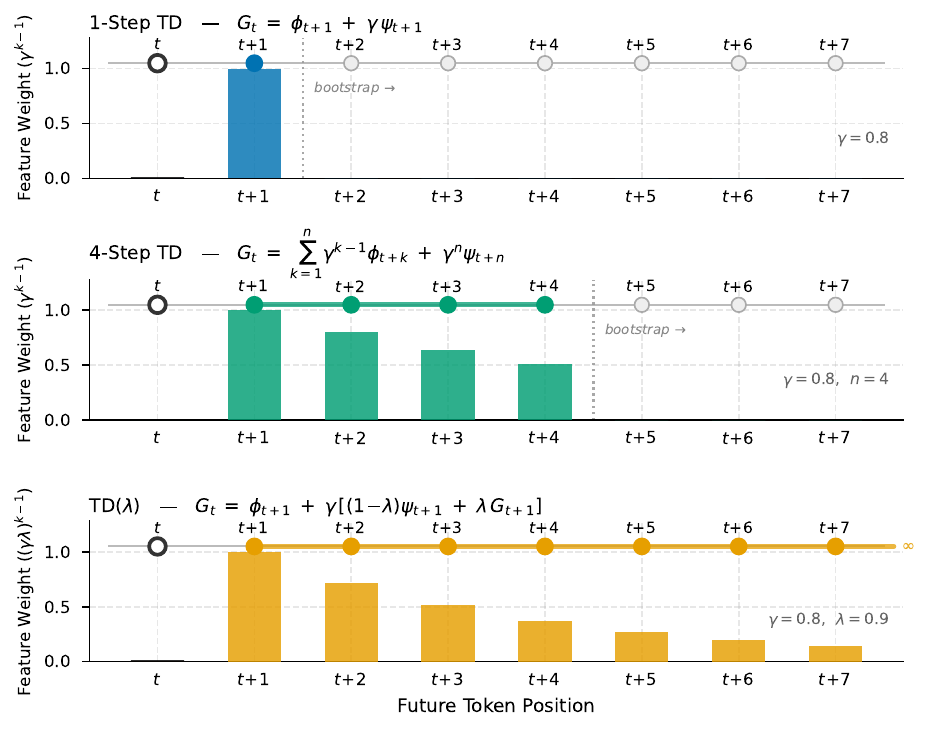}
    \caption{\textbf{TD learning targets for successor representations.} 1-Step TD (blue),  $n$-Step TD shown for $n=4$ (teal), and TD($\lambda$) (gold) differ in how many future token positions contribute discounted features before bootstrapping and how the individual steps are weighted. They range from a single step to an infinite exponentially-weighted horizon.}
    \label{fig:td}
\end{figure}

The dynamics of the environment are typically unknown, and for large state spaces tabular estimates are infeasible, so the \gls{sr} must be learned from samples with a function approximator~\citep{barreto2017successor}, trained on \gls{td} targets~\citep{russek2017predictive} of the kind used in the classical \gls{td} algorithm~\citep{sutton1998reinforcement}. We write $\phi_t$ for the feature representation of the state at time $t$, here a one-hot vector encoding which token is observed, and $\psi_t$ for the model's prediction of the \gls{sr} at that state. Learning proceeds by regressing $\psi_t$ onto a target $G_t$ built from future observations; the target types differ in how far ahead they look before bootstrapping on the model's own prediction.

The simplest is the 1-step Bellman target, which combines the next state's features with the prediction at that state rather than waiting for complete trajectories:
\begin{equation}
    G_t = \phi_{t+1} + \gamma\,\psi_{t+1}.
    \label{eq:onestep}
\end{equation}

Longer horizons can be integrated by accumulating discounted features over $n$ steps before bootstrapping at position $t+n$, and TD($\lambda$) generalises this further, interpolating smoothly between the 1-step target of Equation~\eqref{eq:onestep} ($\lambda = 0$) and full Monte Carlo returns ($\lambda = 1$). Figure~\ref{fig:td} states all three targets and shows how each weights future positions. Since we operate on fixed sequences rather than a stream of online interactions, we compute the $\lambda$-return by a backward pass, initialised at the final position as
\begin{equation}
    G_{T-2} = \phi_{T-1} + \gamma\,\psi_{T-1},
    \label{eq:lambda_return_init}
\end{equation}
and recursed backwards as
\begin{equation}
    G_t = \phi_{t+1} + \gamma\!\left[(1-\lambda)\,\psi_{t+1} + \lambda\, G_{t+1}\right],
    \label{eq:lambda_return}
\end{equation}
with $\lambda = 0.9$ throughout. All bootstrap terms are supplied by the target network introduced below and receive no gradient.

\subsection{Architecture}
\label{sec:architecture}
\begin{figure}
    \centering
    \includegraphics[width=\linewidth]{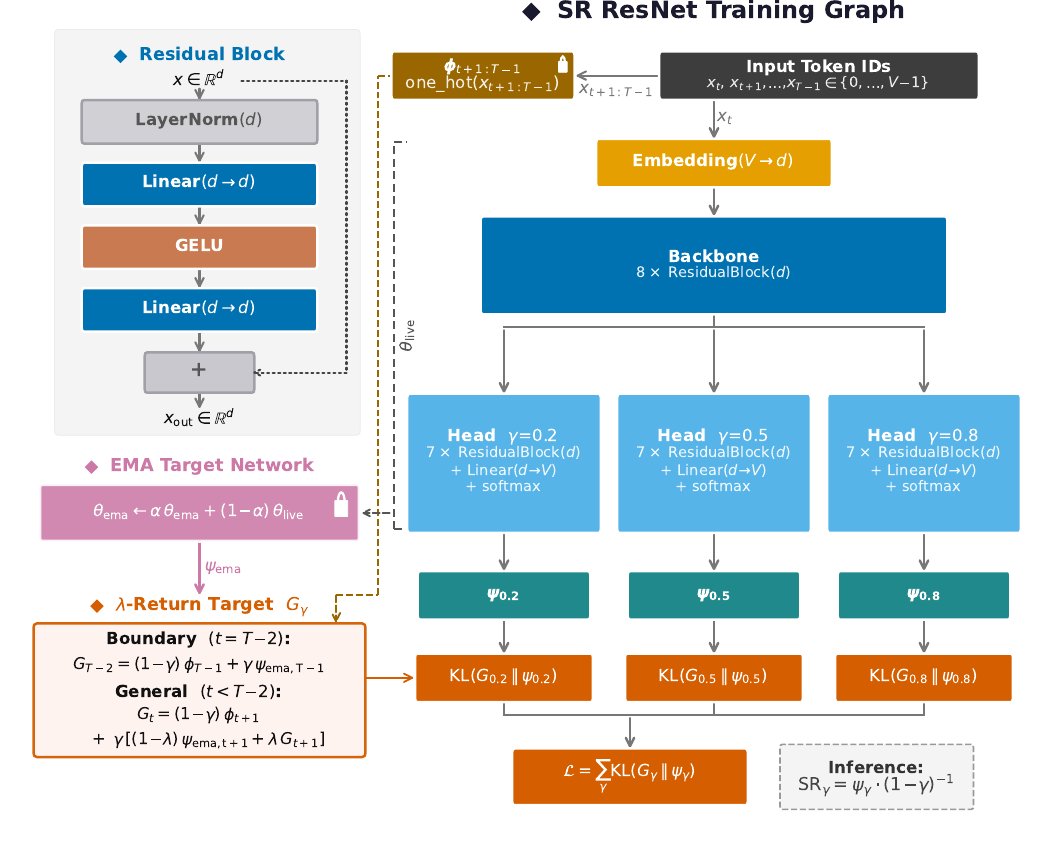}
    \caption{\textbf{Architecture and Training Flow.}
    Input tokens are embedded and passed through a shared backbone that branches into three discount-specific heads, each producing a successor representation prediction. The Residual Block, the main building unit of both backbone and heads, is illustrated top-left. Training targets are computed via a $\lambda$-return formulation that bootstraps from the one-hot encodings of all future tokens and a frozen EMA target network, which maintains a slowly updating copy of the live network weights (lock symbols indicate no gradient flow). The model is optimised by minimising the KL divergence between predictions and targets, summed across all heads. At inference, the true Successor Representation is recovered by a simple rescaling.
    }
    \label{fig:archi}
\end{figure}

Our \gls{sr} model follows the broad residual network paradigm~\citep{he2016deep, he2016identity}, adapted to a fully-connected setting; Figure~\ref{fig:archi} shows the full model.
All components are built from a common building block, the \textit{ResidualBlock}, which applies a two-layer MLP with pre-normalisation and an additive skip connection: LayerNorm, a linear map, a GELU nonlinearity, a second linear map, and addition of the block input. It is drawn in the figure's top-left panel and written out, with its design choices, in Supplementary Appendix~\smBlock{}.

Each token ID is embedded and passed through a shared backbone of these blocks. Three parameter-independent heads are attached to the resulting hidden representation, one per discount factor $\gamma \in \{0.2, 0.5, 0.8\}$, each a further block stack closed by a linear projection and a softmax over the full vocabulary, giving the predicted successor distribution $\psi_\gamma$. Keeping the heads separate lets each specialise for its temporal horizon, mirroring the increasing scale of place fields along the hippocampal longitudinal axis~\citep{stachenfeld2017hippocampus}. Widths and depths are given in Table~\ref{tab:sr_hyperparams}.

The blocks operate independently at each sequence position: no recurrence, attention, or convolution mixes information across positions. Each prediction is therefore conditioned on the current token ID alone. Since a word's continuation depends on more than its identity, distinct contexts collapse onto a single context-averaged estimate per word -- state aliasing. Multi-step targets do not remove it, but they determine what the aliased estimate averages over. A one-step target would leave everything beyond the immediate successor to the model's own averaged predictions, converging on the \gls{sr} implied by the corpus bigram statistics; the $\lambda$-return instead carries the realised continuation: even at the longest horizon, $71\%$ of the discounted target mass comes from observed tokens rather than from the bootstrap (Supplementary Appendix~\smLambda{}).

\subsection{Training}

We train the \gls{sr} model on a lowercased version of WikiText-103~\citep{merity2016pointer}, a $103$M-token corpus of Wikipedia articles, using the spaCy tokeniser~\citep{honnibal2020spacy}, and reuse the corpus' own article-disjoint train--validation split. Each line is tokenised independently and contributes its first 80 tokens as one sequence, padded where shorter, so targets are confined within a line and never span an article boundary. Padding is excluded from the loss as a prediction site but not from the target, where it takes at most $2.4\%$ of the target mass, at the longest horizon (Supplementary Appendix~\smLambda{}).
The vocabulary consists of the 20{,}000 most frequent tokens; all remaining words are mapped to a single generic unknown token, which accounts for approximately 5.5\% of corpus tokens. Seventeen part-of-speech-specific unknown entries from an earlier preprocessing design survive in the released vocabulary as feature dimensions with no probability mass; removing them leaves the reported \gls{pos} agreement unchanged (Supplementary Appendix~\smRobust{}).

To stabilise training, a frozen shadow copy of the model is maintained alongside the live parameters as an exponential moving average with decay $\alpha$, and supplies the bootstrap targets for the TD loss. This prevents the moving-target instability that arises when a model updates its own targets, analogous to the target network in \gls{rl}~\citep{mnih2015human, haarnoja2018soft}.

Rather than a standard \gls{mse} loss, which is unstable for vocabularies of this size, we used \gls{kl} divergence~\citep{kullback1951information}, re-framing the learning objective as matching a probability distribution over future states. Since the unscaled \gls{sr} sums to $\frac{1}{1-\gamma}$, we scale the one-hot feature $\phi_{t+1}$ by $(1-\gamma)$ and bootstrap on the normalised prediction $\psi_{\mathrm{ema}}$ of the target network, so that the $\lambda$-return target $G_t$ of the preceding section sums to $(1-\gamma) + \gamma = 1$ and constitutes a valid distribution over the vocabulary. Normalisation also makes the objective bear on the relative structure of future visitations rather than their absolute magnitude. Each head is then trained against this normalised target,
\begin{equation}
    \mathcal{L}_\gamma = D_{\mathrm{KL}}\!\left(
        G_\gamma \;\Big\|\; \psi_\gamma
    \right),
    \label{eq:loss}
\end{equation}
with the total loss $\mathcal{L} = \sum_\gamma \mathcal{L}_\gamma$ summing over all three heads; in practice the divergence is evaluated in log-space for numerical stability.
Rescaling $\psi_\gamma$ by $\frac{1}{1-\gamma}$ recovers the final \gls{sr}, restoring the expected discounted occupancy interpretation.
All remaining optimisation settings are reported in Table~\ref{tab:sr_hyperparams}.

\begin{table}[htbp]
\centering
\caption{Architecture and training hyperparameters for the Successor Representation model.}
\label{tab:sr_hyperparams}
\begin{tabular}{lc}
\toprule
\textbf{Hyperparameter} & \textbf{Value} \\
\midrule
Vocabulary size         & 20{,}000 \\
Sequence length         & 80 \\
Embedding dimension     & 512 \\
Hidden size             & 512 \\
ResNet blocks (trunk)   & 8 \\
ResNet blocks (heads)   & 7 ($+$ linear projection) \\
Total parameters        & 56{,}283{,}744 \\
Discount factors $\gamma$ & 0.2, 0.5, 0.8 \\
Trace parameter $\lambda$ & 0.9 \\
Batch size              & 160 \\
Learning rate           & $1 \times 10^{-4}$ \\
LR scheduler            & Cosine, decaying to $0$ \\
Warmup steps            & 1{,}000 \\
Gradient clipping       & 1.0 \\
Weight decay            & $1 \times 10^{-5}$ \\
Epochs                  & 10 \\
EMA decay $\alpha$      & 0.999 \\
\bottomrule
\end{tabular}
\end{table}

\subsection{Evaluation Vocabulary and POS Lexicon}
To analyse the structure learned by the \gls{sr} model, we first construct a \gls{pos} lexicon by tagging the first 2{,}000 lines of the WikiText-103 training split with spaCy's large English model (\texttt{en\_core\_web\_lg}) and assigning each word type its majority \gls{pos} tag, resolving lexical ambiguity (e.g., bank as noun vs.\ verb) by majority voting. Words absent from that sample (18 of the 1{,}377 in the extended vocabulary, none in the NVA subset) are tagged with a unigram tagger trained on the Penn Treebank~\citep{marcus1993building}, as implemented in NLTK~\citep{bird2009natural}. These labels enter the analysis only as post-hoc reference annotations, never as clustering inputs.
We then extract \glspl{sr} for up to 200 tokens per part-of-speech category, selecting the most frequent tokens within each category, considering either only nouns, verbs, and adjectives (NVA) or all major \gls{pos} categories, excluding tags: \texttt{X, SYM, SPACE, PUNCT, INTJ}. This led to 600 tokens in the NVA subset and 1{,}377 in the extended vocabulary, distributed across tags as shown in Table~\ref{tab:pos_tags}.
The \gls{sr} vectors analysed throughout span the full 20{,}000-dimensional vocabulary axis, including the reserved entries noted above. They are the heads' logits, equal to the log-transformed \gls{sr} up to one constant per word, the scale conventionally used for distributional statistics in language research~\citep{levy2014neural}: on the probability scale a Euclidean metric is dominated by a few high-probability entries (Supplementary Appendix~\smScale{}). Prior to clustering, the high-dimensional \gls{sr} vectors are denoised and reduced via \gls{pca}~\citep{pearson1901liii,hotelling1933analysis}, retaining 99.99\% of the variance, which suppresses spurious dimensions while preserving the dominant structure.

\begin{table}[htbp]
\centering
\caption{Token counts by \gls{pos} tag.}
\label{tab:pos_tags}
\small
\begin{tabular}{lc}
\toprule
\textbf{\gls{pos} Tag} & \textbf{Token Count} \\
\midrule
ADJ   & 200 \\
VERB  & 200 \\
PROPN & 200 \\
ADV   & 200 \\
NOUN  & 200 \\
NUM   & 200 \\
ADP   & 58  \\
PRON  & 43  \\
AUX   & 27  \\
SCONJ & 22  \\
DET   & 13  \\
CCONJ & 10  \\
PART  & 4   \\
\midrule
\textbf{Total} & \textbf{1{,}377} \\
\bottomrule
\end{tabular}
\end{table}

\subsection{Consensus Clustering}

To obtain groupings that do not rest on a single choice of $K$, we employed consensus clustering by evidence accumulation~\citep{fred2005combining} over 240 trials per algorithm. Such ensembles take their diversity either from resampling the data~\citep{monti2003consensus} or, as here, from varying the resolution of the base clusterings. Each trial clusters at its own base resolution $K_{\text{base}}$, one of the 24 prime numbers from 3 to 97, so that no base resolution is a multiple of another. Each is run ten times under a different random seed, and the co-association matrix therefore reflects structure that persists across scales.

We ran two disjoint sets of trials, one per base algorithm: $K$-Means~\citep{bishop2006pattern}, which favours hard, spherical boundaries, and \glspl{bgm}, which use full covariance matrices and Dirichlet process priors to capture ellipsoidal structure while suppressing components the data do not support; $K_{\text{base}}$ therefore fixes the number of clusters exactly for $K$-Means but acts as an upper bound for the mixtures (Supplementary Appendix~\smBaseClust{}). Each set yields its own consensus solution per target resolution, so agreement between them speaks to the geometry rather than either algorithm's inductive bias.

Within each set, a co-association matrix records how often each pair of tokens is co-assigned across the 240 base clusterings. Average-linkage agglomerative clustering on this matrix yields a hierarchy, and cutting it at a target resolution $K$ gives the reported solutions. The aggregation is itself seeded: re-seeding it reproduces the aggregate agreement metrics and the three-class solution, but not individual memberships or cluster numbering, so the solutions below are one representative partition rather than a unique one (Supplementary Appendix~\smRobust{}). Resolution scans re-cut that same hierarchy at every integer $K$ from 2 to 100 for the NVA subset and from 2 to 200 for the extended vocabulary, while detailed analyses are showcased at $K = 3$ and $K = 30$ (NVA) and $K = 13$ and $K = 120$ (extended).

\subsection{Cluster Evaluation}
Clustering quality is assessed against \gls{pos} tags using three complementary metrics, defined formally in Supplementary Appendix~\smMetrics{}.
Purity, the fraction of tokens in a cluster belonging to its majority class, shows per cluster which groupings align with \gls{pos} tags and which do not, but rises mechanically with $K$ and so cannot be compared across resolutions.
Two global metrics serve that comparison. \Gls{nmi}~\citep{strehl2002cluster} penalises fragmentation but is not adjusted for chance, whereas \gls{ari}~\citep{hubert1985comparing} is, which leaves it approximately resolution-independent and makes it the more reliable basis for comparison across values of $K$.

For visualisation, \gls{sr} vectors are projected into 2D via \gls{umap}~\citep{mcinnes2018umap} ($n_{\text{neighbors}}=15$, $\text{min\_dist}=0.1$). Clusters are coloured by their majority \gls{pos} tag, with lightness and hue variants distinguishing multiple clusters sharing the same category.

\subsection{Transition Networks}
To examine the sequential structure captured by the \gls{sr}, we compute inter-cluster transition weights. Restricting targets to the token set $\mathcal{T}$ entering the clustering, we pool the predicted successor mass from each cluster $C_i$ towards each cluster $C_j$ and normalise each row, giving $M_{ij}$, the share of cluster $i$'s pooled successor mass that falls on cluster $j$. The source cluster's size and the $(1\!-\!\gamma)^{-1}$ rescaling to the \gls{sr} are constant within a row and cancel, so the statistic is unaffected by either (Supplementary Appendix~\smTransitions{}).

All reported weights are conditional on the target lying in $\mathcal{T}$ and therefore characterise
within-$\mathcal{T}$ structure only; because tokens are pooled before normalisation, source tokens whose
successor mass stays within $\mathcal{T}$ contribute proportionally more.
The statistic does not adjust for target-cluster cardinality, target unigram frequency, sampling uncertainty, or multiple comparisons across the $K^2$ edges; it is used as a descriptive summary rather than an inferential enrichment measure.

We visualise these weights as heatmaps and as directed graphs in which each cluster forms a node, edge width is proportional to transition strength, and self loops are omitted. The $K = 3$ graph shows all inter-cluster transitions; the $K = 120$ subnetwork is restricted to the displayed clusters and shows every transition among them above $0.005$, so the number of outgoing edges varies by cluster.

\section{Results}\label{sec3}

\begin{figure}
    \centering
    \includegraphics[width=\linewidth]{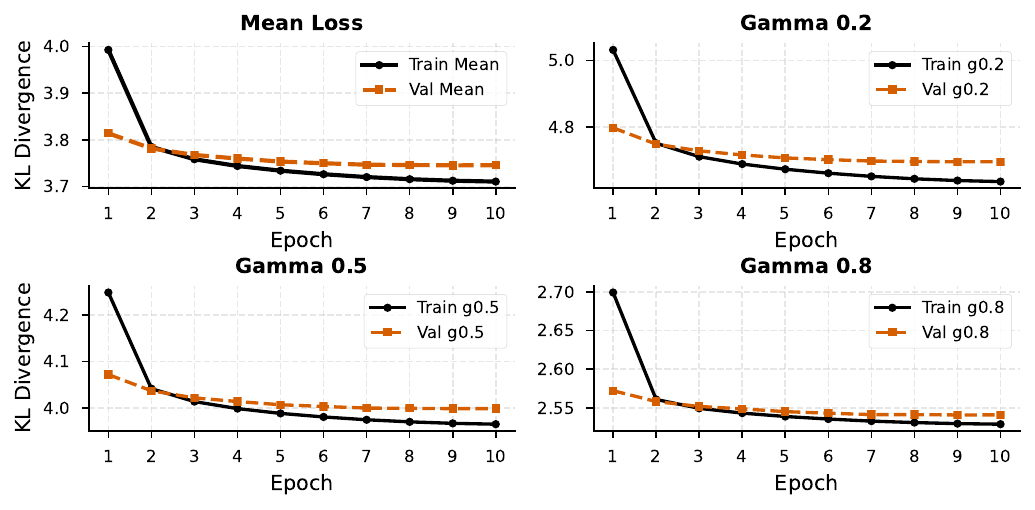}
    \caption{\textbf{Training and Validation Losses.} Mean and per-head training (solid black) and validation (dashed red) losses over 10 epochs.}
    \label{fig:loss_curves}
\end{figure}

\subsection{Training Dynamics and Emergent Embedding Geometry}

Training loss (Figure~\ref{fig:loss_curves}) drops sharply between the first and second epochs and then declines slowly, whereas validation loss falls more gradually; by epoch 10 it has plateaued, changing by less than $0.001$ over the final three epochs, and all subsequent analyses use this checkpoint.
The train--validation gap reaches $0.9\%$ of the loss at epoch 10, indicating no appreciable overfitting.

The heads differ mainly in level, with higher $\gamma$ converging lower: final training losses of $4.64$, $3.97$, and $2.53$ nats for $\gamma = 0.2$, $0.5$, and $0.8$.
Because the targets are normalised distributions, all three heads share a minimum of zero, so the ordering is not a scale artefact.
Nor does it rank the heads: as $\gamma$ grows the targets approach the corpus-level stationary distribution and differ less between tokens, making a low divergence easier to attain.
The detailed cluster analyses that follow accordingly use the $\gamma = 0.2$ head, where that variation is greatest, while the comparisons across horizons cluster each head in turn.

\begin{figure}[t]
    \centering
    \includegraphics[width=\linewidth]{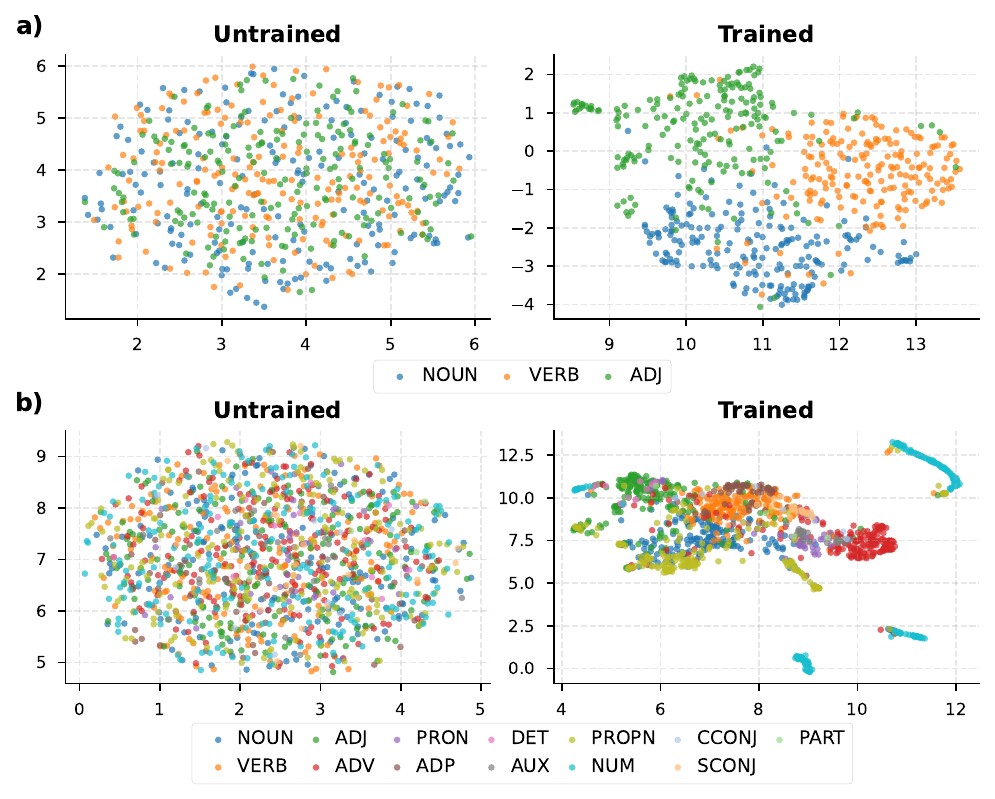}
\caption{\textbf{Emergence of Syntactic Structure in \gls{sr} Geometry.} UMAP 
projections of $\gamma = 0.2$ \gls{sr} embeddings for the untrained model (left) and after 
    training (epoch 10, right), coloured by the spaCy-derived reference \gls{pos} tag. Prior to 
training, embeddings form a structureless cloud with no systematic \gls{pos} 
organisation. \textbf{a)}~NVA subset (NOUN, VERB, ADJ): after training, 
embeddings decompose into three spatially separated regions corresponding to 
the three \gls{pos} categories. \textbf{b)}~Extended vocabulary (13 \gls{pos} 
categories): after training, most categories converge into a central region 
with emergent substructure, while a subset project as isolated peripheral 
strands; numerals show the most distinct separation.}
 \label{fig:structure_emergence}
\end{figure}

What that head has learned is visible in the embedding geometry (Figure~\ref{fig:structure_emergence}).
Before training, embeddings form a structureless cloud; after training, the three NVA categories occupy spatially distinct regions, so the learned \glspl{sr} separate nouns, verbs, and adjectives from successor statistics alone.
The extended vocabulary separates less cleanly: most categories converge into a central region, consistent with their overlapping distributional profiles, whereas numerals form isolated peripheral strands, anticipating the narrow, largely self-contained successor distributions analysed below.
Separation is thus graded rather than uniform, and the following subsections quantify this organisation at increasing cluster resolution.

\begin{figure}
    \centering
    \includegraphics[width=0.96\linewidth]{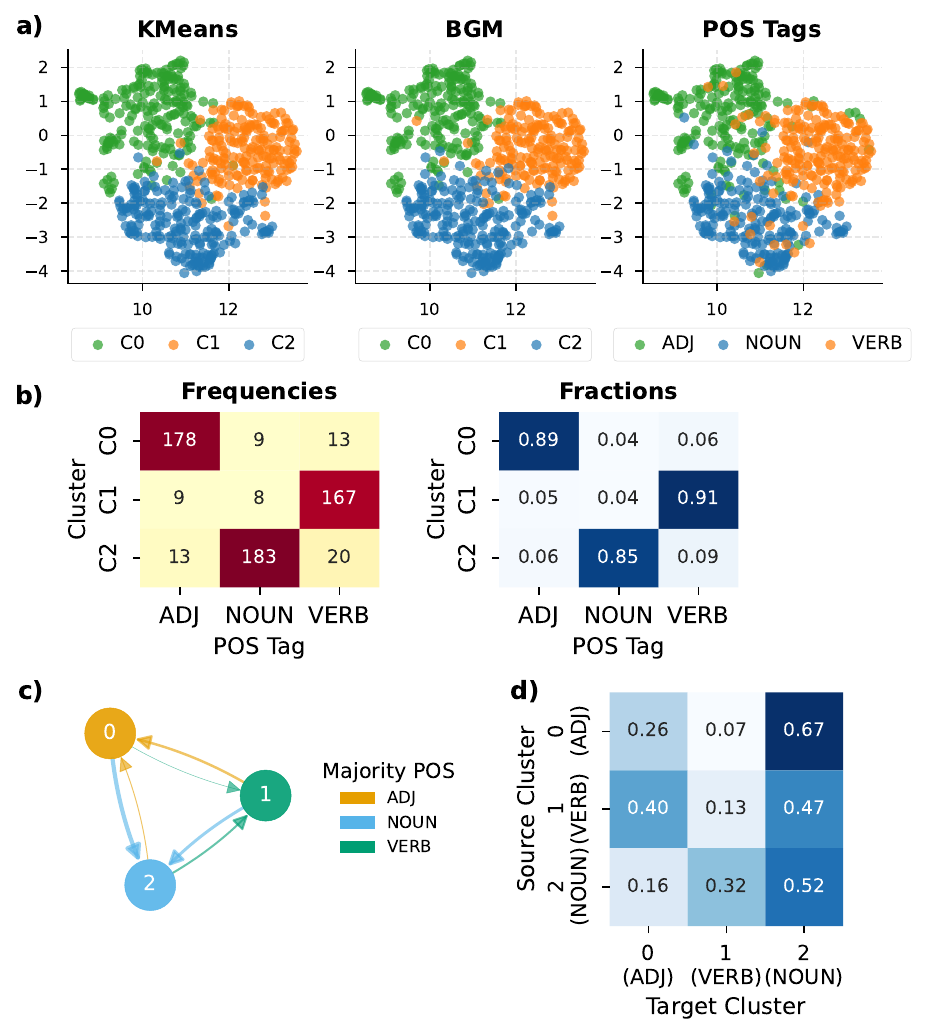}
  \caption{\textbf{Unsupervised Clustering of Successor Representations and Connectivity Structure (NVA, $K=3$).} To assess the geometric structure of the main syntactic categories in embedding space, \glspl{sr} with $\gamma=0.2$ of the 200 most frequent tokens were sampled from each of three part-of-speech categories  (nouns, verbs, and adjectives), yielding a balanced dataset of 600 tokens.
    \textbf{a)}~UMAP projections of token embeddings, coloured by KMeans consensus cluster assignment (left),
    BGM consensus cluster assignment (centre), and spaCy-derived reference
    \gls{pos} tags (right), enabling direct visual comparison between
    unsupervised cluster solutions and known syntactic categories.
    \textbf{b)}~Cluster purity heatmaps for KMeans consensus clusters, showing the raw \gls{pos} tag
    frequency per cluster (left) and the fractional \gls{pos} tag composition per cluster
    (right), quantifying the degree to which each cluster captures a coherent
    syntactic category.
    \textbf{c)}~Directed inter-cluster \gls{sr} connectivity network.
    Nodes represent KMeans consensus clusters and are coloured according to their majority \gls{pos} tag;
    directed edges indicate the magnitude of \gls{sr} connectivity from source to target cluster, with edge thickness proportional to connection strength.
    \textbf{d)}~\gls{sr} connectivity matrix corresponding to the network in~(c).
    Rows denote source clusters and columns denote target clusters; each entry reports
    the fraction of total directed \gls{sr} connectivity from the source to the target cluster.}
    \label{fig:clustering_sr_nva_3}
\end{figure}

\subsection{Recovering Basic Syntactic Categories}

Consensus clustering at $K = 3$ on the NVA embeddings ($\gamma = 0.2$, the 200 most frequent tokens per category) recovers the three syntactic categories.
In the colour-coded \gls{umap} projections of Figure~\ref{fig:clustering_sr_nva_3}a the two consensus solutions both align with the spaCy-derived reference \gls{pos} tags and agree with each other on token assignment (\gls{ari} $0.93$), so the partition belongs to the learned geometry rather than to either clustering construction (full token-level assignments are provided in Supplementary Appendix~\smNVAthree{}).
Scored against those labels, the KMeans-derived solution reaches per-cluster purities between $0.85$ and $0.91$ across three near-balanced clusters, one per category (Figure~\ref{fig:clustering_sr_nva_3}b).

Beyond cluster identity, the inter-cluster \gls{sr} network and connectivity matrix (Figures~\ref{fig:clustering_sr_nva_3}c,d) show mass asymmetries compatible with familiar syntactic transitions.
Adjectives are the clearest case: 67\% of their mass goes to nouns and only 7\% to verbs, matching the adjective--noun modifier relation.
Nouns divide theirs between nouns and verbs, consistent with nominal compounding and subject--predicate structure, while verbs divide between nouns and adjectives and retain little internally.
That low verb-internal share is partly a matter of the tag set: Universal Dependencies assigns auxiliaries a separate \texttt{AUX} tag, placing auxiliary-to-main-verb transitions outside the verb cluster.
Nouns, however, are the largest target for every source cluster.
The asymmetries between rows --- adjective-to-verb transitions are near-absent, while the reverse direction carries $40\%$ of verb mass --- are therefore more informative than the absolute shares.
Whether the space is organised only at this coarse level is taken up next.

\begin{figure}
    \centering
    \includegraphics[width=0.94\linewidth]{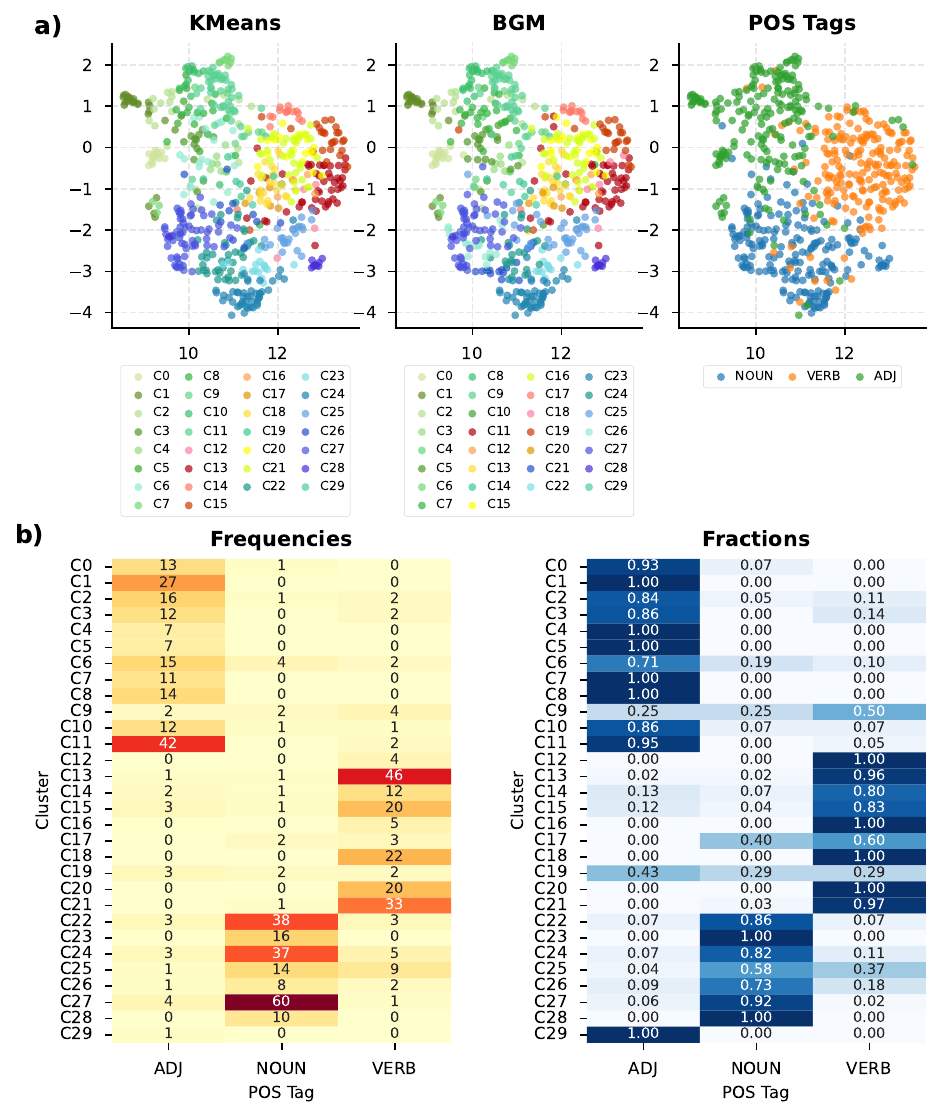}
     \caption{\textbf{Unsupervised Clustering of Successor Representations (NVA, $K=30$).}
    \Glspl{sr} for $\gamma=0.2$ were extracted for a balanced sample of the 200 most frequent tokens per
    part-of-speech category (nouns, verbs, and adjectives), yielding a dataset
    of 600 tokens in total.
    \textbf{a)}~UMAP projections of token embeddings, coloured by KMeans-derived consensus cluster
    assignment (left), BGM-derived consensus cluster assignment (centre),
    and spaCy-derived reference \gls{pos} tags (right). Clusters are aligned across algorithms by maximising cluster overlap. They receive a base colour for their joint majority \gls{pos} tag, with distinct shades differentiating individual clusters within the same syntactic category to facilitate comparison.
    \textbf{b)}~Cluster purity heatmaps evaluating the alignment between KMeans-derived consensus
    clusters and \gls{pos} categories. The left heatmap shows the raw \gls{pos} tag frequency across the 30 clusters, and the right heatmap shows the fractional composition of each cluster,
    i.e.\ the proportion of a cluster's tokens belonging to each \gls{pos} category, whose maximum
    is that cluster's purity, illustrating the degree of syntactic specialisation achieved at
    higher cluster granularity ($K=30$).}
    \label{fig:clustering_nva_30}
\end{figure}

\subsection{Fine-Grained Clustering: Lexical Sub-Classes}

At $K = 30$ the same geometry resolves into a mosaic of localised sub-regions while the three-way \gls{pos} topology of the $K = 3$ solution is preserved (Figure~\ref{fig:clustering_nva_30}a, coloured by KMeans-derived and BGM-derived assignment and by the spaCy-derived reference \gls{pos} labels).
Most sub-clusters remain dominated by a single category: 24 of 30 reach a purity of at least $0.8$, and 12 reach $1.0$ (Figure~\ref{fig:clustering_nva_30}b).
Nine of the 30 clusters are described below; full token-level assignments for all of them are listed in Supplementary Appendix~\smNVAthirty{}.

The sub-clusters are also semantically coherent.
Adjectives separate into ordinal and rank modifiers (\textit{first, second, final}; C0), size and extent terms (\textit{large, small, entire}; C5), and colour and non-gradable properties (\textit{white, black, female}; C4).
Verbs split between a broad eventive class (\textit{built, came, returned}; C13, the largest of the verb clusters) and two narrower ones: epistemic and communicative verbs (\textit{said, stated, suggested}; C18) and competitive outcome verbs (\textit{won, win, scored}; C12).
Nouns group by domain, including measurement units (\textit{km, ft, mph}; C28) and cultural and media artefacts (\textit{film, album, episode}; C24).

Not all of the remaining mixing is noise.
The largest of the six clusters that fall below $0.8$, C25 (purity $0.58$), is systematically mixed: it collects temporal nouns (\textit{time, year, night}) together with phase verbs (\textit{began, started, ended}).
Grouping of this kind --- by semantic-functional domain rather than by syntactic category --- recurs in the extended vocabulary below.

The following subsection turns to a separate, quantitative criterion: \gls{pos} agreement across \gls{sr} horizons and cluster granularities.

\begin{figure}
    \centering
    \includegraphics[width=\linewidth]{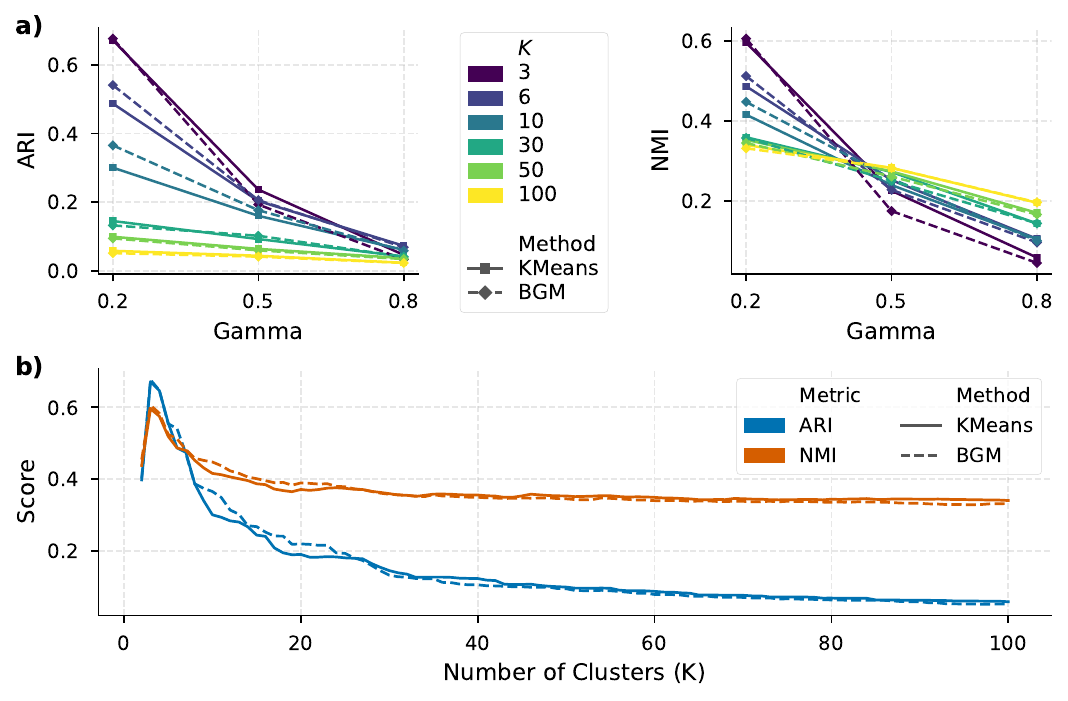}
\caption{\textbf{Clustering quality metrics (ARI and NMI) across \gls{sr} configurations for the NVA subset.}
Both metrics are invariant to label permutation, but only \gls{ari} is adjusted for chance.
    \textbf{a)}~\gls{ari} (left) and \gls{nmi} (right) as a function of the discount factor $\gamma$, evaluated on the NVA subset. Trends are shown for varying cluster granularities ($K$), comparing KMeans-derived (solid lines) and BGM-derived (dashed lines) consensus solutions.
    \textbf{b)}~\gls{ari} and \gls{nmi} as a function of cluster count $K$ at fixed $\gamma = 0.2$, illustrating how cluster resolution affects the recovery of the reference syntactic categories from the NVA subset.}
    \label{fig:ari_nmi_nva}
\end{figure}

\subsection{Clustering Quality in the NVA Subset}

Figure~\ref{fig:ari_nmi_nva} scores both consensus solutions against the spaCy-derived reference \gls{pos} labels across a grid of $\gamma$ and $K$.
Both \gls{ari} and \gls{nmi} decrease monotonically as $\gamma$ increases, for every $K$ shown and both algorithms (Figure~\ref{fig:ari_nmi_nva}a).
\Gls{pos} information is therefore most distinct when immediate successors carry the greatest weight, and the separation declines as longer trajectories contribute more.

At $\gamma = 0.2$ both metrics peak sharply at $K = 3$ (\gls{ari} at $0.67$ and \gls{nmi} at $0.60$) for both consensus solutions, matching the number of \gls{pos} categories in the subset (Figure~\ref{fig:ari_nmi_nva}b).
Beyond that peak they diverge.
\Gls{ari} falls steeply, to below $0.06$ at $K = 100$, whereas \gls{nmi}, which is not chance-adjusted, declines gradually and settles near $0.34$.
The asymmetry follows from how the two measures treat over-segmentation: \gls{ari} penalises splitting a category across clusters, whereas \gls{nmi} is much less sensitive to it, since internally homogeneous sub-clusters remain informative about the label.
The two consensus solutions track each other across the grid, so this pattern is not specific to either method.

Whether the same organisation holds beyond three categories is taken up next.

\begin{figure}
    \centering
    \includegraphics[width=\linewidth]{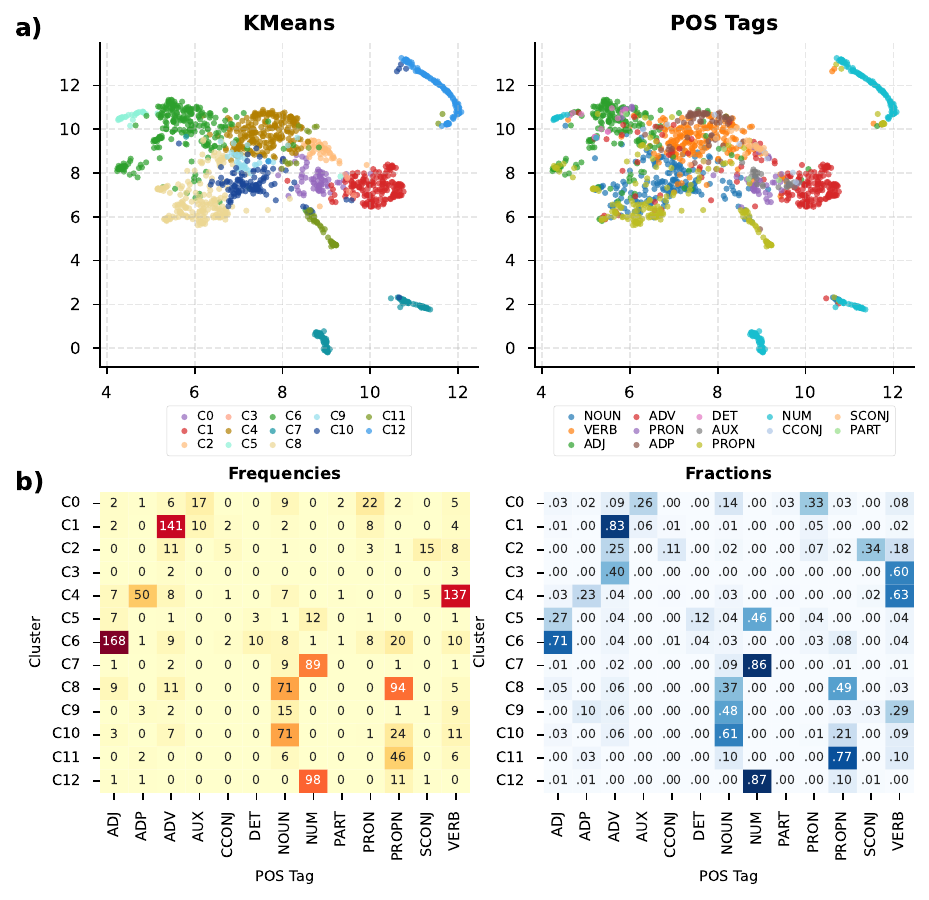}
    \caption{\textbf{Clustering of Successor Representations (extended vocabulary, $K=13$).}
    \Glspl{sr} for $\gamma=0.2$ were extracted for up to 200 of the most frequent tokens per part-of-speech category across 13 categories, yielding a dataset of 1{,}377 tokens in total.
    \textbf{a)}~UMAP projections of extended-vocabulary token embeddings, coloured by KMeans-derived consensus
    cluster assignment (left) and \gls{pos} tags (right).
    \textbf{b)}~Cluster purity heatmaps evaluating the alignment between the 13 KMeans-derived consensus
    clusters and \gls{pos} categories.}
    \label{fig:clustering_full_k13}
\end{figure}

\subsection{Extended Vocabulary Clustering at 13 Clusters}

Extending to the broader tag set (1{,}377 tokens, 13 \gls{pos} categories, $\gamma = 0.2$, $K = 13$) tests whether the geometry survives a more heterogeneous vocabulary.
It largely does: the \gls{umap} projections in Figure~\ref{fig:clustering_full_k13}a remain broadly organised, with several clusters occupying distinct regions that correspond to the spaCy-derived reference \gls{pos} labels.

Five clusters stay close to single categories, with purities from $0.71$ to $0.88$ (Figure~\ref{fig:clustering_full_k13}b): adverbs (C1), adjectives (C6), proper nouns (C11), and numerals in two separate clusters (C7 and C12); the remaining eight fall from $0.63$ to $0.33$.
Within those categories they carve distinctions the tag set does not, separating bare cardinals and measurement units (\textit{35, km, million}; C7) from calendar years and month names (\textit{1960, April, November}, with \textit{since} and \textit{until}; C12), and confining C11 to titles and the first names that accompany them (\textit{captain, John, Paul}).

Two kinds of mixing recur among the remaining clusters (full token listings in Supplementary Appendix~\smFullThirteen{}).
The first groups items sharing a slot or an adjacency: quantificational modifiers stand with spelled-out numerals in the same pre-nominal position (\textit{several, many}; \textit{two, three}; C5), verbs stand with the prepositions and particles that follow them (a quarter of C4), and clause-introducing material --- subordinators, coordinators, and discourse adverbs (\textit{that, because, however}) --- stands with the reporting verbs that select clausal complements (\textit{said, claimed}; C2).
The second groups items by semantic domain irrespective of category: common and proper nouns merge into a geographic and institutional cluster, with more proper nouns than common ones (\textit{area, city, England}; C8), and into an event and media cluster carrying performer and award names (\textit{film, album, Smith}; C10).

C9 repeats a pattern already seen in the NVA subset.
It combines duration and interval nouns (\textit{days, months, years}) with aspectual verbs (\textit{began, started, ended}), as C25 did at $K = 30$ in the NVA subset, and adds the temporal prepositions (\textit{before, after}) that the NVA subset excludes.
That the same cross-category temporal grouping appears in two subsets at two resolutions points to organisation by semantic-functional domain rather than by syntactic category alone.

The following subsection turns to the finer constructional resolution, focusing on the numerical and temporal subnetwork.

\begin{figure}
    \centering
    \includegraphics[width=\linewidth]{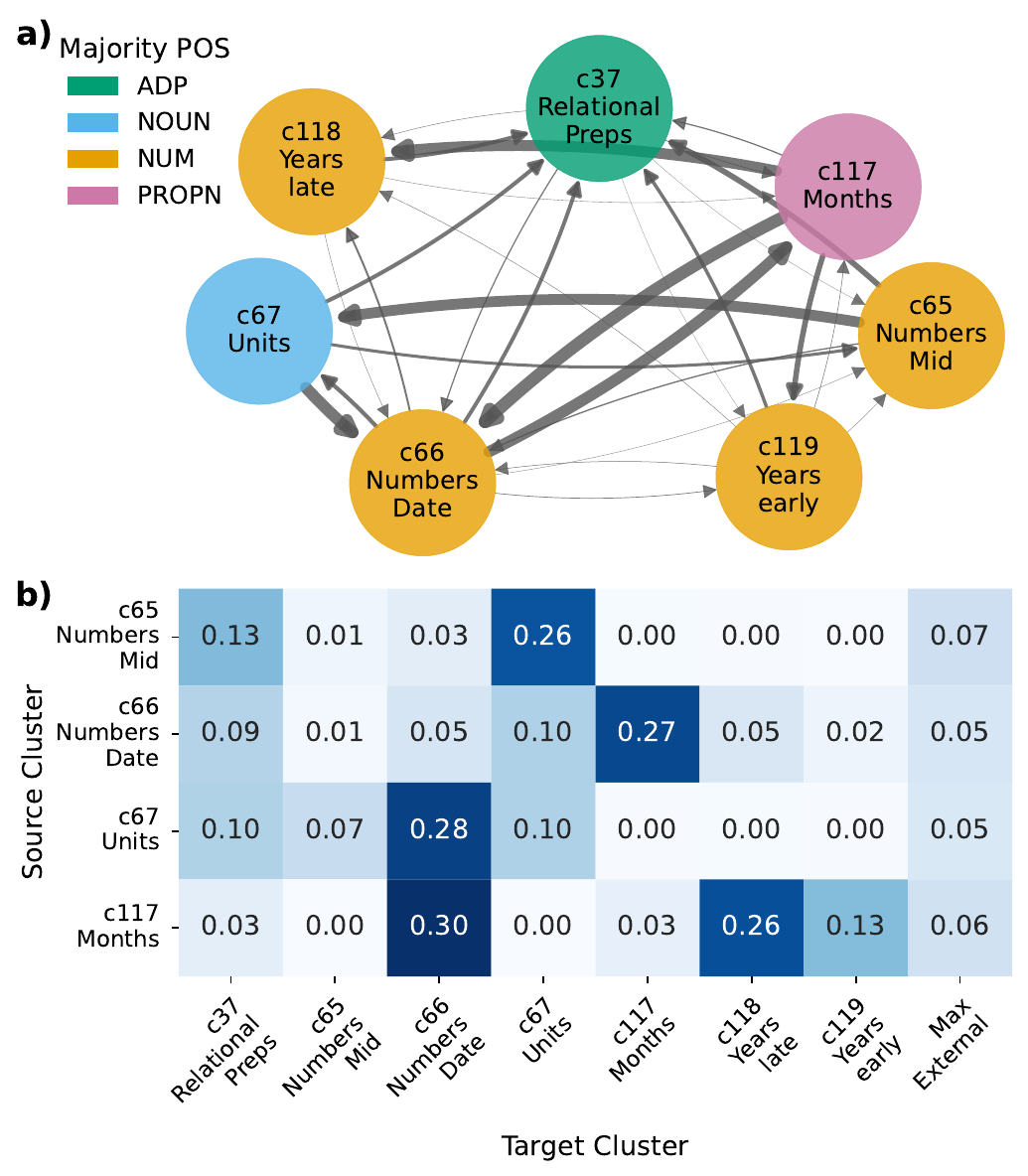}
\caption{\textbf{Successor representation connectivity within numerical and temporal cluster subnetworks.}
    All weights come from the $\gamma=0.2$ head of the $K=120$ extended-vocabulary solution.
    \textbf{a)}~Directed \gls{sr} subnetwork restricted to a selection of numerical and temporal clusters. Nodes represent individual KMeans-derived consensus clusters coloured by their majority \gls{pos} tag; directed edges represent inter-cluster transition strengths.
    \textbf{b)}~\gls{sr} connectivity matrix corresponding to the subnetwork in~(a). The final column (\textit{Max External}) reports the maximum connectivity from each source cluster to any cluster outside the displayed subset.}
    \label{fig:sr_flow}
\end{figure}

\subsection{Cluster Content and Transition Structure at High Resolution}

Figure~\ref{fig:sr_flow} presents a directed \gls{sr} subnetwork of seven clusters from the $K=120$ solution: relational prepositions (C37), measurement units (C67), month names together with a few temporal function words (C117: \textit{until}, \textit{since}, \textit{late}), and four clusters of numerals (C65, C66, C118, C119).
The subnetwork is largely self-contained: the cardinal, unit and month-name clusters each send four to six times more mass to their strongest partner within the set than to any single cluster outside it (\textit{Max External}, Figure~\ref{fig:sr_flow}b).

The four numeral clusters are divided by content, and the divisions are calendrical.
Two of them hold years, one earlier (1910--1965) and one later (1958--2016); of the remaining two, one holds the integers $0$--$31$ and the other those from $32$ upward, placing the boundary exactly at the largest day of the month.
The transitions follow that division.
Day-range cardinals and month names form a near-reciprocal pair, each sending its strongest edge to the other ($0.27$ and $0.30$), and month names pass on to the year clusters.
Numerals above the calendar range instead feed measurement units, which feed back into the day-range cardinals, while their outflow toward months and years is negligible.
The subnetwork therefore separates two constructions that draw on the same numerals, date expressions and measurement phrases, and routes each through its own sequence.

Comparable specialisation appears elsewhere in the $K=120$ solution: modal and negation markers (C0: \textit{not, would, can, must}) are separated from copular and aspectual auxiliaries (C7: \textit{was, is, were, been}), and person-role titles (C116: \textit{captain, director, king}) direct their strongest outflow to relational prepositions and their second strongest to personal given names (C115) --- the two continuations a title actually takes.
Those titles and given names had formed a single cluster at $K = 13$ (C11), so across the two resolutions the same relation is carried first by shared cluster membership and then by a directed edge.
The complete $K=120$ token assignments are provided in Supplementary Appendix~\smFullOneTwenty{}, and these transition patterns together identify candidate sequential relationships within complex expressions such as date phrases, measurement constructions, and title--name sequences.

\begin{figure}
    \centering
    \includegraphics[width=\linewidth]{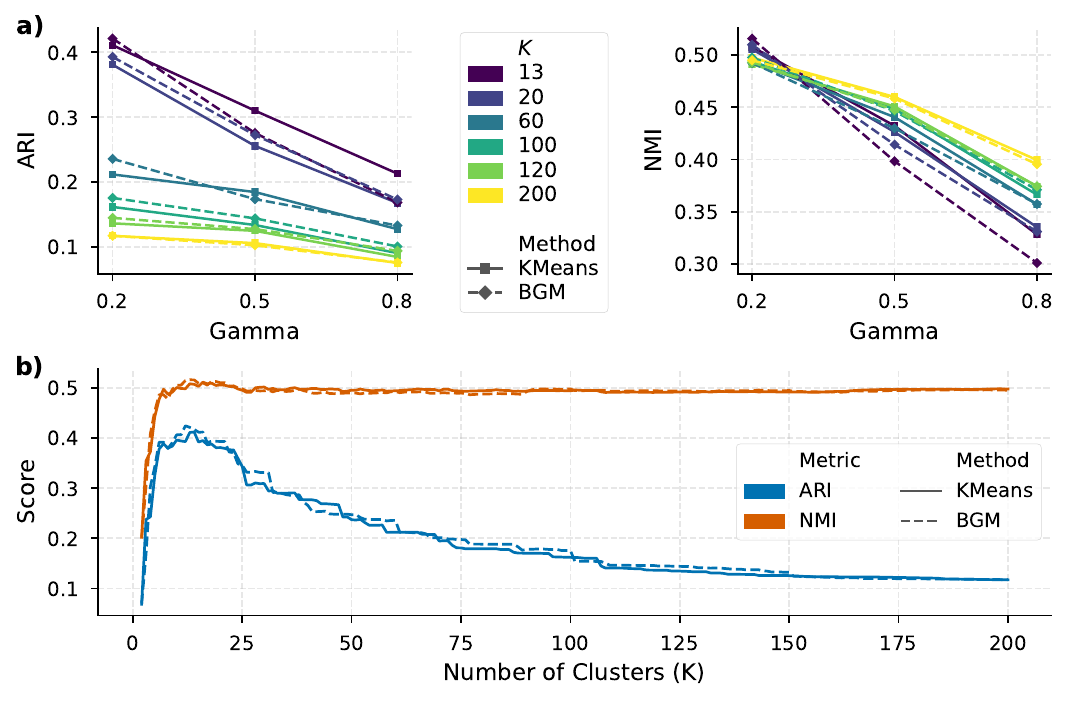}
    \caption{\textbf{Clustering quality metrics (\gls{ari} and \gls{nmi}) across \gls{sr} configurations for the extended set of \gls{pos} tags.}
    \textbf{a)}~\gls{ari} (left) and \gls{nmi} (right) as a function of the discount factor $\gamma$, evaluated on the extended \gls{pos} tag set. Trends are shown for varying cluster granularities ($K$), comparing KMeans-derived (solid lines) and BGM-derived (dashed lines) consensus solutions.
    \textbf{b)}~\gls{ari} and \gls{nmi} as a function of cluster count $K$ at fixed $\gamma = 0.2$, illustrating how cluster resolution affects the recovery of syntactic structure across the extended vocabulary.}
    \label{fig:ari_nmi_full}
\end{figure}

\subsection{Clustering Quality in the Extended Vocabulary}

The same \gls{ari} and \gls{nmi} analysis applied to the extended vocabulary (Figure~\ref{fig:ari_nmi_full}) reproduces the $\gamma$ trend but not the $K$ trend.
Both metrics again decrease monotonically with $\gamma$, for every $K$ shown and both algorithm-derived consensus solutions (Figure~\ref{fig:ari_nmi_full}a).
Recovery of the spaCy-derived reference labels is therefore again strongest when immediate successors receive the greatest weight.

At fixed $\gamma = 0.2$, however, the two metrics dissociate as $K$ increases (Figure~\ref{fig:ari_nmi_full}b), where in the NVA subset they had declined together.
\Gls{ari} peaks near $0.41$ at $K \approx 13$ --- the number of \gls{pos} categories present --- and then declines steadily to $0.12$ at $K = 200$.
\Gls{nmi} peaks in the same range, at just above $0.51$, but does not follow that decline, remaining between $0.49$ and $0.50$ across the remainder of the scan.
The dissociation is the same asymmetry seen in the NVA subset, in stronger form: there \gls{nmi} still declined gradually, whereas here it does not decline at all.
The two consensus solutions track each other closely throughout, indicating that this behaviour is not specific to either method.

The theoretical implications of this emergent structure, and its relationship to constructionist accounts of word classes and grammatical knowledge, are taken up in the Discussion.

\section{Discussion}\label{sec4}

\subsection{Emergent Word Classes and Their Reach} %

The central result is that neural \glspl{sr} trained directly on a large natural-language corpus develop a geometry that contains recoverable word-class and candidate construction-like structure. At the shortest horizon ($\gamma=0.2$), clustering the 200 most frequent nouns, verbs, and adjectives per class recovers the three categories with per-cluster purities of 0.85--0.91. The directed mass between those clusters is asymmetric in ways familiar from syntax: 67\% of adjective outflow reaches nouns.

The result connects two lines of previous work. It extends emergent category structure from next-word prediction and its extension to two-word POS combinations~\citep{surendra2023word,krauss2024word} to an explicitly multi-step predictive objective. It also extends neural \gls{sr} models from restricted and artificial domains~\citep{stoewer2022neural,stoewer2023neural,stoewer2023conceptual,stoewer2025multi} to natural language, analysed at the level of individual word types. The claim is one of sufficiency: multi-step prediction makes the structure recoverable.

The same categories recur across objectives and architectures, the kind of convergence attributed to models approximating the same underlying reality~\citep{huh2024platonic}. But what varies in that convergence is the model, not the language. The earlier replication did span two languages, though German and English are closely related and share much of the typological profile on which the present method relies. Whether familiar word classes constitute language-independent universal categories remains contested, particularly for adjectives. The criteria identifying word classes are themselves language-particular~\citep{haspelmath2010comparative}; in many languages, property words share their predication construction with verbs~\citep{croft2001radical}, and some languages have been analysed as lacking a distinct adjective class or even a clear noun--verb distinction~\citep{evans2009myth}. This does not preclude convergence at a more abstract relational level: similar distributional or predictive geometries could recur across languages without necessarily yielding the same discrete category boundaries. The informative question is therefore how far such structures extend across languages, which aspects converge, and where they break down---questions that distributional analyses of individual languages can address directly.

Within English, the three categories have coherent signatures in future-word distributions. This is compatible with usage-based and constructionist accounts, in which category membership reflects recurrent positional and combinatorial privileges rather than only an intrinsic lexical feature~\citep[cf.][]{bybee2010language,hopper1987emergent,goldberg2006constructions,diessel2019usage}.

\subsection{Finer Resolutions Reveal Construction-Like Structure}

The extended-vocabulary analysis shows why \gls{pos} recovery is not the only relevant criterion: the finer the resolution, the more the partition tracks recurrent environments rather than categories. At $K=13$ several clusters cross \gls{pos} boundaries while remaining interpretable: reporting verbs, for instance, stand with clause-introducing material such as \textit{that} and \textit{because} (C2). The clearest case is C9, which groups duration nouns with aspectual verbs and temporal prepositions. Its noun--verb core recurs as C25 in the NVA subset at a different resolution, the prepositions appearing only in the extended vocabulary. That the grouping appears in two subsets at two resolutions makes it unlikely to be an artefact of either.

The numerical subnetwork shows how fine that organisation becomes. At $K=120$ the geometry separates date from measurement uses of the same numerals, a division no \gls{pos} tag encodes. Day-range cardinals and month names feed each other, month names pass on to the year clusters, and numerals above the range feed measurement units instead. That boundary is calendrical: the split falls at 31, the largest day of the month.

Because a word is represented by what follows it and nothing else, grouping by successor profile draws together sets of two kinds. Paradigmatic sets have members substitutable in one slot (\textit{several, many} with \textit{two, three}; C5). Syntagmatic pairs have members that are sequentially related rather than substitutable (verbs with the prepositions that follow them; C4). Both kinds occur at $K=13$, where syntagmatic co-membership is common, but it is rare among the $K=120$ clusters examined. There the clusters are internally homogeneous, and the sequential relation is carried instead by the edges between them. Person-role titles and the given names they precede show the shift directly: at $K=13$ they share a cluster, and at $K=120$ they are separate clusters joined by a directed edge.

Every cluster groups items by shared successor profiles; what differs is whether the tag set has a name for the region they occupy. Named regions look like formal categories; the unnamed ones like semantic domains (C9), shared slots (C5), or sequential pairs (C4), and are not an artefact of fine partitioning: count-based bigram vectors at the same resolution subdivide within the reference categories (Supplementary Appendix~\smBaseline{}). The objective favours none, so the unnamed clusters are not automatically classification errors; we describe them as \emph{construction-like}. They are corpus-derived candidates requiring validation against independent construction annotations, behavioural data, or neural measurements---done at the sentence level for argument-structure constructions~\citep{ramezani2026convergent}, though not yet for type-level clusters.

This reading has a theoretical anchor in the characterisation of constructions as clusters of lossy memory traces~\citep{goldberg2019explain}. \Gls{sr} vectors provide a computational counterpart: each compresses a word's distributional history into an expected future profile. The cluster count $K$ then indexes the grain of the compression, though nothing measured here identifies a psychological threshold.

Seen in a longer historical perspective, the present results extend a progression in what distributional learning has been shown to recover. Self-organising maps first demonstrated that contextual statistics can arrange words into topographic semantic and grammatical categories \citep{ritter1989self}; recurrent next-word prediction then recovered lexical classes and their subclasses, and subsequently richer grammatical structure \citep{elman1990finding,elman1991distributed}. More recent neural prediction studies reproduced word-class organisation in natural language and extended it to two-word category combinations \citep{surendra2023word,krauss2024word}, while successor representations recast such organisation as regions of a predictive map, initially in a small controlled language \citep{stoewer2022neural}. The present results bring these lines together at natural-language scale: the same predictive geometry contains broad word classes, finer semantic-functional groupings, and candidate construction-like relations whose sequential structure can become explicit in the directed transitions between regions. At the sentence level, related representational convergence has already been observed for argument-structure constructions \citep{ramezani2026convergent}. Across these studies, the common principle is therefore not a particular architecture, but the emergence of linguistic categories and structure from relational or predictive context rather than from category labels supplied in advance.

\subsection{A Memory Mechanism for Constructional Knowledge}

If constructions are clusters of memory traces, the system that forms them is in the first instance a memory system. The question becomes whether representations of this form are learnable by a plausible memory mechanism. Each step of that path has independent support. The hippocampus underpins episodic and relational memory~\citep{eichenbaum2000cortical} and supports cognitive maps~\citep[cf.][]{tolman1948cognitive} of spatial and non-spatial structure. \Glspl{sr} have been proposed as the computational form of those maps because they encode expected future occupancy rather than only current position~\citep{stachenfeld2017hippocampus}. Human neuroimaging reports \gls{sr}-consistent signals~\citep{garvert2017map,brunec2022predictive,ekman2023successor}, and TD($\lambda$) learns such representations with computational, synaptic, and behavioural backing~\citep{russek2017predictive,bono2023learning,kahn2025trial}. The present model, however, is trained on the TD($\lambda$) target under a \gls{kl} objective, not by the incremental updates those studies describe.

Just as place cells compress trajectories of spatial experience into a predictive map of an environment, token-level \glspl{sr} compress trajectories of linguistic experience into a geometry reflecting the combinatorial structure of the language. Constructional knowledge may then arise from a general predictive memory system operating over structured sequences rather than from a specialised linguistic module. What the results establish is computational compatibility rather than biological implementation: architecture, objective, and data scale are engineering choices, not a process-level model of human learning (see Limitations and Future Directions). The heads were analysed independently and never combined, though the multi-head design is compatible with representations operating over several timescales at once~\citep{momennejad2018predicting}.

Multiscale representation itself is a recurring organisational principle of the brain. Along the hippocampal longitudinal axis, the spatial scale of place-cell representations increases from fine dorsal/posterior to progressively coarser ventral/anterior representations, paralleled by an increase in grid spacing along the dorsoventral medial entorhinal cortex~\citep{brun2008progressive}. Recent high-resolution fMRI extends this principle beyond spatial representation: during predictable sequences, hippocampal activity followed a continuous long-axis gradient consistent with coarser anterior and finer posterior coding, while violations of learned structure additionally recruited content-sensitive subregions for semantic, spatial, and combined mismatches~\citep{jordan2026spatial}. This complements evidence that grid-like coding generalises from physical to abstract conceptual spaces~\citep{constantinescu2016organizing}. Multiscale organisation is also evident at the cortical level: intrinsic timescales increase from sensory towards association and prefrontal cortex~\citep{murray2014hierarchy}, while the rostrocaudal prefrontal axis has been associated with increasingly abstract levels of hierarchical control~\citep{badre2009rostro}. Together, these findings suggest that the brain represents structured experience simultaneously across multiple spatial, temporal, and abstract scales. The present SR heads should not be mapped directly onto particular brain regions, but their different predictive horizons instantiate the same general computational principle: representing future structure at multiple scales in parallel~\citep{momennejad2018predicting}.

\subsection{Short Horizons Carry the Word Classes}

The discount factor $\gamma$ is fixed before training and shapes the representation. Agreement with the reference \gls{pos} labels is highest at the shortest horizon and falls as $\gamma$ increases, across both subsets, both consensus solutions, and every cluster count shown. The information separating the categories is therefore concentrated in the nearest successors. What organises the geometry at longer horizons is left open. Increasing $\gamma$ promoted coarse taxonomic clustering in earlier \gls{sr} maps of handcrafted semantic features~\citep{stoewer2023neural}. A count-based extension recovers semantic concepts from natural language at $\gamma = 0.9$~\citep{haga2025unified}, against the $\gamma = 0.2$ at which agreement peaks here.

The cluster count $K$ is imposed after training and only partitions the result; no value of it is theoretically privileged. Agreement peaks near the number of reference categories, which shows those categories are recoverable, not that the geometry has no finer structure. Beyond that peak the two metrics dissociate in the extended vocabulary, \gls{ari} declining steadily while \gls{nmi} does not. Because \gls{ari} penalises splitting, that decline records over-segmentation relative to the tag set rather than uninformative clusters. Neither curve can show whether the finer divisions are linguistically meaningful; the evidence for that is qualitative as in the numeral and title divisions at $K=120$.

Two properties of the procedure bear on how the resulting clusters should be read. The procedure weights every following position by distance alone, so no single environment decides a word's category; linear order, though, is the only relation it sees. It also turns two analyst decisions into parameters: how far the evidence extends, and how coarsely the result is partitioned. A finding can be reported across their range, making its dependence on either choice visible. The procedure does not ask whether a language has a given word class, but how sharply the corresponding region separates, and under which horizon and resolution.

\subsection{Limitations and Future Directions}

The analysis reports a single training run, and two further models trained under different seeds reproduce its agreement with the reference labels and its dependence on $\gamma$ and $K$, but not its exact partition. They match the reported partitions at $K = 3$ and $K = 13$ at an \gls{ari} of $0.86$ to $0.93$, as closely as re-seeding the consensus aggregation on the reported model does. At the finer resolutions they match at $0.67$ to $0.80$, somewhat below that (Supplementary Appendix~\smRobust{}). The clustered features also carry a per-word constant that the \gls{kl} objective does not determine; clustering the successor distribution in logarithmic coordinates that discard it recovers the same categories at least as strongly (Supplementary Appendix~\smScale{}). Cluster-level descriptions therefore characterise one representative solution rather than a unique partition, and groupings recurring across subsets and resolutions weigh correspondingly more.

Out-of-vocabulary words are collapsed onto a single unknown token whose frequency cutoff is itself correlated with \gls{pos}; removing the reserved dimensions from the clustering features leaves agreement with the reference labels essentially unchanged (Supplementary Appendix~\smRobust{}). Subword modelling would avoid that collapse altogether, but remains untested, as does the geometry's organisation of types rarer than those evaluated here. The model learns one representation per lowercased word type, conflating distinct senses and grammatical uses: \textit{may} as modal or proper name, \textit{since} as temporal or causal. That is a general limitation of type-level distributional representations~\citep{camacho2018word,fonteyn2021varying}, and it bears particularly on clusters read as constructional slots. Contextualised \glspl{sr}, conditioned on preceding context rather than averaged across occurrences, would separate those token-specific roles.

\Gls{pos} tags never enter optimisation, but they shape the evaluation: they select a balanced three-class subset, reduce each word to its majority tag, and then score the clusters against those same three classes. Purity and \gls{nmi} are adjusted for neither chance nor class size, so neither is interpretable alone in the imbalanced extended vocabulary; the chance-adjusted \gls{ari} is reported alongside them and carries the comparisons across $K$. The cluster-transition statistic likewise aggregates predicted mass without adjusting for cluster size or marginal token frequency, and no edge is tested against an inferential null, so the directed patterns remain descriptive pending exposure-adjusted, held-out analysis.

No comparison model was trained on the same corpus and vocabulary, so superiority over next-token language modelling is not established; a matched one-step model would settle it. Specificity to the \gls{sr} objective is addressed by untrained controls: bigram conditionals, count-based co-occurrence vectors, and the analytic \gls{sr} computed in closed form, all clustered by the same procedure (Supplementary Appendix~\smBaseline{}). At the resolutions matching the reference categories none reproduces the reported recovery as reliably: the closed-form \gls{sr} approaches it on the NVA subset, though not stably across seeds, and falls well short on the extended vocabulary.

The $\gamma$ contrast is not a pure horizon manipulation. With $\lambda$ held fixed, the observed-token share of the target falls from 98\% to 71\% as $\gamma$ rises (Supplementary Appendix~\smLambda{}), so horizon and target composition vary together. Isolating the horizon would need $\lambda = 1$ at every head, where the bootstrap term vanishes and the heads differ only in $\gamma$, at the cost of noisier targets and of truncation at the line boundary that no bootstrap absorbs. Either way, \gls{pos} agreement registers only what is lost, not what replaces it; measuring that would need semantic similarity benchmarks, thematic-role and discourse-relation annotations, or held-out prediction of construction membership.

WikiText-103 is large but homogeneous, and some of the most interpretable structures---calendar dates and institutional titles---reflect its encyclopaedic register. Conclusions about language acquisition or English grammar generally would need conversational, literary, and child-directed corpora, and cross-linguistic comparison would need register-matched ones, unavailable for most languages. Another language would add limits of its own. The state is the orthographic word, so morphologically rich languages need morphemes as units, with the segmentation deciding what the recovered categories are categories of. Where word order is freer, linear successor statistics carry less of the categorial signal, so a failure to separate could reflect the method rather than the language.

No behavioural or neural data are analysed, so the hippocampal and cognitive-map interpretation advanced above remains a motivated hypothesis rather than a validated neurocomputational account. Nor is the training procedure a model of acquisition: the results concern what is learnable from distributional experience, not how a child learns it. Representational similarity analysis could test whether \gls{sr} geometry predicts neural similarity during continuous language processing~\citep{kriegeskorte2008representational}, and priming, expectation, or production experiments could test the directed transition structure. The model is also ungrounded: earlier \gls{sr} models already associated linguistic and visual information~\citep{stoewer2025multi}, whereas this one isolates sequential structure, and combining the two would test whether construction-like clusters survive perceptual and situational grounding.

\section{Conclusion}\label{sec5}

This study connects two earlier observations by training multi-horizon \glspl{sr} directly on natural language: local next-word prediction can induce word-class information, and neural \glspl{sr} can form predictive maps in restricted domains. The learned geometry recovers major word classes, finer lexical subclasses, and category-crossing groupings, with no \gls{pos} classification objective. The directed mass between them reproduces familiar grammatical asymmetries, and at finer resolutions traces candidate date, measurement, and title--name constructions. \Gls{pos} agreement declines as the horizon lengthens.

All of it follows from future-word distributions alone, on a single encyclopaedic corpus and at the level of word types; the agreement measures survive independent re-training, though the exact cluster memberships do not, and how far any of it extends across languages and registers remains open. \Glspl{sr} have also been proposed as hippocampal cognitive maps---a general memory mechanism, not a language-specific one---and while the claim here stays computational rather than biological, it makes that geometry a testable target for human data. Thus, \glspl{sr} offer a framework for investigating how local transitions and longer predictive horizons jointly shape linguistic representation, providing a concrete example of how categorical linguistic structure can arise as geometry in a predictive state space.

\backmatter

\section*{Declarations}
\addcontentsline{toc}{section}{Declarations}

\ifanon\else
\bmhead{Funding}
This work was funded by the Deutsche Forschungsgemeinschaft (DFG, German Research Foundation): grants KR\,5148/3-1 (project number 510395418), KR\,5148/5-1 (project number 542747151), KR\,5148/10-1 (project number 563909707), and GRK\,2839 (project number 468527017) to PK, and grants SCHI\,1482/3-1 (project number 451810794) and SCHI\,1482/6-1 (project number 563909707) to AS.
\ifmergesupp\else%
For the publication fee, the authors acknowledge financial support by Heidelberg University.
\fi
\fi

\bmhead{Competing interests}
The authors declare no competing interests.

\bmhead{Ethics approval and consent to participate}
Not applicable. The study analyses a publicly available text corpus and involves no human participants or animals.

\bmhead{Consent for publication}
Not applicable.

\bmhead{Data availability}
The WikiText-103 corpus analysed in this study is publicly available~\citep{merity2016pointer}. The trained model, the successor-representation vectors, all cluster assignments and the results behind every reported statistic are deposited at Zenodo (\ifanon available to reviewers at \url{\reviewDataURL}; DOI withheld for anonymous review\else DOI: \href{https://doi.org/\zenodoDOI}{\zenodoDOI}\fi)\ifanon\else~\cite{immertreu2026data}\fi; the deposit includes a script that recomputes and checks sixty-two of the reported values.

\bmhead{Code availability}
The code for every step of the analysis, from corpus preprocessing and training through the clustering, the supplementary analyses and all figures, is available at \ifanon\url{\reviewCodeURL}\else\url{\codeURL}, archived at Zenodo (DOI: \href{https://doi.org/\codeDOI}{\codeDOI})~\cite{immertreu2026code}\fi. Its tests recompute the reported values from the deposited data. Regenerating the clusterings in a different numerical environment yields equivalent but not identical partitions.

\ifanon\else
\bmhead{Author contributions}
Conceptualization: Mathis Immertreu, Achim Schilling, Patrick Krauss. Methodology: Mathis Immertreu, Patrick Krauss. Software: Mathis Immertreu. Formal analysis: Mathis Immertreu. Investigation: Mathis Immertreu. Data curation: Mathis Immertreu. Validation: Mathis Immertreu. Visualization: Mathis Immertreu. Writing -- original draft: Mathis Immertreu, Achim Schilling, Patrick Krauss. Writing -- review and editing: Mathis Immertreu, Achim Schilling, Thomas Kinfe, Patrick Krauss. Supervision: Achim Schilling, Patrick Krauss. Project administration: Patrick Krauss. Funding acquisition: Achim Schilling, Patrick Krauss. Resources: Thomas Kinfe, Patrick Krauss. All authors read and approved the final manuscript.
\fi

\bmhead{Use of generative AI}
Generative AI tools were used during the preparation of this manuscript under the direction and control of the authors. The tools used were OpenAI Codex (GPT-5.5 and GPT-5.6); Claude Code, GitHub Copilot, and Google Antigravity running Claude Sonnet and Claude Opus models (versions 4.5–5) and Claude Fable 5; and Google Gemini (versions 3 and 3.1, Flash and Pro). These tools were used for exploring implementation options, software coding assistance, analysis and figure-rendering code, reference management, LaTeX preparation, copyediting, and improving the clarity and organisation of author-written text. Generative AI was not used to generate research data or scientific results, nor to make substantive scientific decisions. All AI-assisted code and manuscript content were reviewed and verified by the authors, who take full responsibility for the work.

\ifmergesupp
\clearpage
\renewcommand{\thetable}{S\arabic{table}}
\setcounter{table}{0}
\renewcommand{\theHtable}{S\arabic{table}}%

\section*{Supplementary Appendix}

\begin{appendices}

\section{NVA Cluster Assignments at Three Clusters}
\label{app:nva_3}
\noindent Configuration: $\gamma=0.2$, $k=3$.
\textbf{Cluster 0}: same, new, use, own, particular, other, s, known, separate, such, form, individual, level, associated, similar, current, first, possible, key, important, related, limited, whole, previous, larger, additional, real, original, different, single, standard, active, available, common, future, minor, remaining, successful, traditional, independent, create, certain, regular, full, subsequent, notable, potential, popular, third, complete, main, specific, range, local, unknown, recent, various, number, true, major, strong, powerful, clear, significant, entire, wide, special, second, modern, poor, overall, multiple, numerous, primary, prominent, several, high, american, mixed, official, complex, open, european, smaller, contemporary, century, last, commercial, fourth, older, critical, using, free, many, early, difficult, international, top, initial, famous, direct, national, chinese, small, living, good, great, final, greatest, next, secret, large, right, best, long, higher, fifth, canadian, better, few, indian, largest, provide, historical, russian, white, german, professional, japanese, french, daily, body, opening, 20th, increased, private, heavy, short, dead, low, italian, soviet, light, seventh, total, annual, female, highest, sixth, foreign, black, military, double, average, polish, bad, natural, 19th, hard, big, positive, roman, central, federal, greater, latin, western, personal, cultural, ancient, middle, public, native, male, anti, human, dark, armed, social, civil, political, legal, economic, medical, physical, nuclear, senior, lower, severe, imperial, yellow, southern, religious, naval, electric, maximum, estimated, upper, financial, 1st

\textbf{Cluster 1}: created, removed, followed, addition, worked, included, made, taken, put, done, introduced, given, found, seen, allowed, added, used, see, revealed, saw, involved, including, making, return, served, claimed, performed, returned, held, suggested, support, offered, appeared, reported, brought, called, said, recorded, established, led, way, compared, wanted, stated, take, described, noted, turned, came, placed, discovered, believed, formed, produced, leave, shows, ran, released, raised, took, developed, leaving, wrote, run, changed, carried, taking, announced, make, give, present, gave, opened, following, supported, designed, saying, selected, moved, order, ordered, replaced, inspired, keep, find, allow, thought, based, required, lost, pass, takes, sent, left, awarded, provided, signed, include, working, featured, come, running, includes, launched, composed, intended, played, passed, appear, built, playing, considered, planned, issued, went, directed, appears, lead, hit, met, going, caused, proposed, leading, published, fell, responsible, decided, go, sold, features, member, forced, moving, get, do, destroyed, arrived, received, joined, failed, named, know, captured, extended, agreed, won, written, killed, help, according, able, felt, win, told, runs, entered, asked, old, helped, winning, former, became, become, appointed, praised, got, reached, damaged, remains, unable, prime, ranked, becoming, nearby, located, remained, scored, defeated, born, t, suffered, fellow, married

\textbf{Cluster 2}: time, work, place, events, days, position, part, event, group, result, life, line, history, years, day, end, site, company, games, success, role, development, period, project, list, set, case, months, example, release, show, hours, area, weeks, series, point, performance, works, rest, center, times, region, version, year, beginning, office, house, members, play, studio, men, production, start, game, man, stage, country, town, force, action, track, system, field, story, band, campaign, latter, month, power, scene, building, side, film, station, team, post, book, shot, career, war, base, island, record, week, season, service, minutes, tour, control, others, half, death, match, world, party, points, school, bridge, design, children, song, crew, score, command, magazine, home, cast, attack, people, title, head, players, review, night, women, album, construction, fire, division, style, songs, started, ended, name, characters, episode, ground, areas, completed, battle, writing, victory, family, coast, age, forces, ship, front, cut, training, recording, state, damage, music, broadcast, fight, sound, road, player, class, having, art, chart, close, m, ships, live, father, offensive, likely, c, began, radio, government, guns, football, college, land, summer, love, character, square, operation, finished, continued, population, h, league, aircraft, air, least, fighting, media, law, feet, video, son, mother, much, police, fleet, sea, championship, television, species, troops, re, died, northwest, storm, km, late, wife, miles, route, president, spent, ft, water, critics, comic, rear, mm, mph, musical, hurricane

\newpage

\section{NVA Cluster Assignments at 30 Clusters}
\label{app:nva_30}
\noindent Configuration: $\gamma=0.2$, $k=30$.
\textbf{Cluster 0}: fourth, third, fifth, second, seventh, sixth, first, next, last, final, 20th, 19th, opening, 1st

\textbf{Cluster 1}: russian, french, german, italian, chinese, polish, american, soviet, canadian, indian, japanese, european, local, independent, national, international, foreign, military, roman, civil, federal, imperial, professional, naval, private, senior, armed

\textbf{Cluster 2}: known, modern, prominent, century, traditional, contemporary, older, popular, famous, native, unknown, ancient, powerful, living, western, latin, central, southern, anti

\textbf{Cluster 3}: higher, high, low, average, increased, total, lower, highest, overall, maximum, greater, estimated, middle, upper

\textbf{Cluster 4}: white, black, dead, male, female, yellow, dark

\textbf{Cluster 5}: larger, large, small, smaller, largest, entire, whole

\textbf{Cluster 6}: level, single, s, standard, wide, long, range, open, full, short, complete, right, light, double, body, hard, top, free, heavy, big, electric

\textbf{Cluster 7}: political, economic, social, cultural, legal, religious, historical, financial, personal, public, medical

\textbf{Cluster 8}: previous, subsequent, recent, new, future, current, original, successful, official, regular, initial, early, annual, daily

\textbf{Cluster 9}: use, using, create, form, such, associated, provide, available

\textbf{Cluster 10}: other, several, various, numerous, many, multiple, few, certain, individual, different, separate, additional, remaining, number

\textbf{Cluster 11}: particular, similar, same, possible, important, specific, related, key, own, significant, real, common, potential, limited, minor, primary, strong, critical, poor, difficult, true, clear, good, active, notable, main, complex, major, direct, special, greatest, better, mixed, commercial, positive, great, best, bad, natural, secret, physical, human, severe, nuclear

\textbf{Cluster 12}: won, win, winning, scored

\textbf{Cluster 13}: removed, ran, moved, held, returned, formed, opened, came, carried, raised, served, return, run, built, left, leave, destroyed, leaving, placed, fell, turned, reported, arrived, established, passed, went, lost, pass, launched, come, running, captured, sold, hit, moving, going, runs, issued, located, reached, go, damaged, extended, entered, caused, killed, nearby, suffered

\textbf{Cluster 14}: intended, decided, wanted, able, order, allowed, unable, failed, required, agreed, planned, forced, help, helped, proposed

\textbf{Cluster 15}: led, joined, replaced, named, met, selected, signed, sent, appointed, supported, member, awarded, according, told, asked, ordered, former, defeated, leading, fellow, old, born, married, ranked

\textbf{Cluster 16}: become, became, becoming, remained, remains

\textbf{Cluster 17}: get, got, t, way, do

\textbf{Cluster 18}: said, stated, revealed, believed, thought, saying, found, shows, suggested, noted, seen, done, see, appeared, claimed, discovered, appears, felt, considered, know, appear, announced

\textbf{Cluster 19}: addition, support, involved, responsible, present, lead, prime

\textbf{Cluster 20}: take, give, making, given, put, taking, made, offered, make, took, gave, taken, takes, brought, find, saw, provided, allow, keep, received

\textbf{Cluster 21}: created, included, introduced, followed, performed, produced, including, described, featured, worked, added, used, recorded, inspired, compared, composed, includes, wrote, called, developed, include, released, based, designed, features, published, directed, following, changed, working, playing, written, played, praised

\textbf{Cluster 22}: work, members, life, case, result, role, group, example, history, others, children, works, men, times, period, man, project, list, people, characters, women, development, death, latter, party, players, father, name, family, war, world, son, mother, likely, age, having, fight, player, law, wife, much, critics, president, re

\textbf{Cluster 23}: event, games, match, season, events, team, success, tour, career, campaign, championship, victory, points, league, football, action

\textbf{Cluster 24}: performance, show, film, release, series, game, book, version, song, story, album, play, writing, music, songs, studio, episode, scene, production, band, magazine, video, recording, stage, track, cast, style, design, title, record, review, shot, score, sound, broadcast, character, art, television, love, live, comic, radio, media, chart, musical

\textbf{Cluster 25}: days, months, weeks, beginning, years, hours, month, time, day, year, start, started, week, ended, minutes, completed, began, night, finished, continued, summer, died, spent, late

\textbf{Cluster 26}: end, place, part, point, set, position, half, post, head, cut, close

\textbf{Cluster 27}: line, area, site, company, force, region, island, center, rest, building, station, town, base, field, bridge, system, service, ground, house, ship, forces, power, areas, construction, office, command, division, control, coast, fire, attack, side, country, front, crew, ships, land, training, battle, damage, road, aircraft, home, state, school, air, fleet, guns, government, sea, troops, class, population, offensive, square, operation, police, northwest, college, fighting, water, species, storm, rear, hurricane

\textbf{Cluster 28}: feet, km, ft, mph, m, h, miles, c, mm, route

\textbf{Cluster 29}: least

\section{Extended Vocabulary: 13 Clusters}
\label{app:full_13}
\noindent Configuration: $\gamma=0.2$, $k=13$.

\textbf{Cluster 0}: them, else, him, myself, you, ourselves, t, should, 'd, here, yourself, people, everything, will, we, forever, characters, not, would, anything, me, herself, might, someone, men, shall, man, do, there, does, something, did, anyone, i, players, children, must, could, just, whoever, 'll, appears, nothing, 're, women, get, can, everyone, got, appear, really, ve, somebody, ll, god, am, able, love, exactly, anybody, may, ta, everybody, ago, hasn, unable

\textbf{Cluster 1}: also, actually, itself, themselves, indeed, therefore, apparently, even, sometimes, yet, be, nevertheless, nonetheless, likewise, consequently, often, initially, occasionally, it, simultaneously, otherwise, instead, already, probably, merely, others, possibly, simply, rather, himself, frequently, thus, been, still, well, being, certainly, once, were, rarely, have, then, had, essentially, either, again, partly, reportedly, is, previously, always, usually, they, perhaps, similarly, was, likely, are, having, subsequently, ultimately, he, who, mostly, traditionally, commonly, eventually, regularly, never, has, generally, soon, solely, easily, normally, typically, now, entirely, largely, specifically, effectively, consistently, quickly, she, increasingly, finally, originally, naturally, properly, remains, equally, literally, clearly, ever, potentially, truly, recently, primarily, mainly, longer, directly, personally, deliberately, immediately, suddenly, partially, completely, rapidly, extensively, briefly, constantly, successfully, permanently, currently, historically, too, fully, virtually, quite, somewhat, significantly, seriously, much, physically, almost, widely, temporarily, repeatedly, far, closely, heavily, poorly, badly, slowly, officially, gradually, slightly, considerably, exclusively, critically, thereby, species, strongly, greatly, more, severely, less, unusually, remained, re, publicly, locally, politically, commercially, fairly, nearly, predominantly, formally, very, positively, highly, actively, extremely, steadily, deeply, accidentally, relatively, unsuccessfully, pretty

\textbf{Cluster 2}: that, though, however, because, but, although, furthermore, where, if, said, and, believed, whether, when, why, wherein, thought, saying, hence, which, additionally, whenever, notably, neither, nor, unless, stated, while, afterwards, how, both, so, what, whilst, whom, meanwhile, afterward, claimed, felt, name, especially, know, particularly, young

\textbf{Cluster 3}: become, became, becoming, formerly, newly

\textbf{Cluster 4}: out, up, apart, of, for, created, followed, down, removed, as, on, included, aside, with, addition, from, including, introduced, to, put, except, part, place, over, found, involved, made, used, seen, than, called, see, added, making, allowed, given, saw, taken, through, back, off, behind, worked, return, compared, away, result, support, suggested, done, established, revealed, returned, described, example, recorded, brought, held, into, discovered, reported, led, formed, placed, in, whereas, served, include, performed, developed, raised, leave, take, by, known, leaving, produced, noted, offered, following, about, supported, turned, includes, along, present, upon, taking, carried, make, based, give, took, appeared, allow, replaced, keep, opened, ran, released, towards, gave, wanted, came, around, like, besides, alongside, changed, moved, inspired, designed, onto, beyond, find, left, ordered, unlike, lost, selected, toward, wrote, required, running, provided, featured, e.g., at, outside, awarded, within, sent, takes, launched, built, due, working, caused, above, composed, signed, intended, considered, proposed, between, planned, responsible, passed, notwithstanding, under, amongst, inside, come, issued, using, moving, destroyed, prior, played, during, playing, among, hit, leading, against, closer, directed, despite, forced, captured, published, across, extended, joined, via, features, went, versus, fell, named, arrived, entered, close, sold, going, received, go, old, according, failed, below, decided, killed, help, former, reached, written, continued, throughout, agreed, nearby, damaged, near, helped, praised, aboard, without, located, ranked, amidst, defeated, opposite, suffered, amid, fellow, vs.

\textbf{Cluster 5}: two, three, five, four, six, seven, several, eight, nine, few, twelve, ten, many, other, twenty, all, numerous, thirty, number, various, remaining, some, multiple, these, those, most

\textbf{Cluster 6}: 's, same, new, own, s, use, the, particular, or, form, whose, such, one, earlier, first, their, individual, separate, level, any, current, similar, associated, possible, kind, another, whole, key, only, single, real, previous, limited, no, original, important, its, plus, related, standard, his, additional, larger, successful, future, active, full, third, common, available, create, different, complete, our, regular, minor, independent, traditional, subsequent, popular, true, clear, second, enough, potential, unknown, strong, certain, powerful, cut, special, notable, range, recent, entire, poor, local, main, overall, her, specific, wide, this, whatever, american, official, last, major, open, fourth, high, significant, modern, century, primary, mixed, complex, every, european, prominent, free, good, english, contemporary, critical, regardless, commercial, difficult, older, worldwide, famous, final, international, great, early, initial, next, top, chinese, direct, each, right, national, smaller, secret, living, long, greatest, fifth, best, small, australian, better, canadian, british, a, white, large, russian, french, african, german, japanese, higher, body, indian, non, dead, provide, professional, opening, 20th, daily, spanish, chief, short, historical, largest, italian, private, increased, little, seventh, low, heavy, soviet, prime, female, black, light, sixth, pre, annual, greek, double, per, bad, total, highest, my, your, military, polish, further, average, overseas, big, an, hard, foreign, 19th, green, grand, positive, roman, personal, natural, latin, federal, greater, public, male, dark, anti, cultural, ancient, human, native, blue, civil, red, armed, deep, musical, legal, political, social, allied, christian, medical, economic, brown, senior, physical, imperial, nuclear, severe, yellow, lower, super, religious, royal, naval, electric, hot, maximum, financial, 1st

\textbf{Cluster 7}: 56, 35, 44, 57, 39, 32, 20, 62, 85, 15, 33, 65, 16, 22, 48, 25, 43, 13, 58, 42, 66, 9, 14, 6, 12, 45, 8, 41, 46, 70, 26, 60, 10, 53, 80, 28, 30, 23, 75, 51, 17, 27, 24, 55, 54, 36, 7, 37, 76, 5, 90, 52, 21, 95, 34, 38, 40, 50, 11, 19, 18, 47, 29, 4, 110, 31, 49, 3, 2, 120, 250, 64, 150, 1, 400, 200, 300, 100, 800, 700, 600, 500, 0, 000, approximately, c, million, hundred, feet, thousand, km, least, ft, billion, m, h, mm, mi, °, miles, mph, estimated, roughly

\textbf{Cluster 8}: area, city, africa, line, canada, company, region, island, europe, town, india, site, wales, australia, america, center, britain, position, territory, england, scotland, china, france, rest, park, group, hall, force, islands, coast, station, country, germany, house, service, kingdom, mexico, run, society, bridge, bank, elsewhere, building, states, cross, point, empire, office, hill, field, division, japan, base, association, ii, army, areas, florida, airport, republic, west, system, command, forces, bay, state, side, district, war, united, north, department, corps, south, london, texas, abroad, school, york, set, union, pass, construction, forth, power, drive, east, council, party, california, museum, street, home, virginia, control, ship, ground, campaign, road, club, county, attack, front, iii, river, university, washington, academy, railway, land, training, battle, fire, church, government, half, head, annually, crew, manchester, congress, pacific, college, uk, brigade, regiment, damage, infantry, runs, ships, northwest, navy, chicago, family, creek, atlantic, angeles, court, michigan, battalion, ocean, squadron, class, boston, st., fleet, sea, parliament, air, machine, troops, square, population, act, lake, aircraft, northeast, oxford, guns, forward, us, police, operation, eastward, los, lead, offensive, northward, studios, ashore, law, fighting, san, depression, route, mountain, highway, water, storm, southern, western, central, offshore, eastern, middle, northern, rear, aft, upper, hurricane

\textbf{Cluster 9}: days, months, weeks, years, day, beginning, hours, time, month, year, start, ended, end, week, started, before, after, minutes, completed, ahead, announced, began, night, shortly, till, finished, summer, died, spent, mid, en

\textbf{Cluster 10}: work, games, together, event, performance, events, show, series, release, success, version, life, history, play, game, role, list, alone, records, film, respectively, times, works, project, case, book, shows, studio, development, story, hers, band, jones, way, song, scene, smith, production, stage, magazine, songs, period, cast, episode, album, record, action, track, career, tour, title, members, season, writing, order, onwards, match, team, review, score, u.s., music, shot, soundtrack, race, award, awards, design, post, johnson, recording, latter, world, williams, style, twice, death, jackson, broadcast, points, d, later, star, grant, video, character, box, player, art, sound, fight, victory, thereafter, win, x, chart, media, screen, live, football, radio, age, television, league, entertainment, cup, conference, rock, b, championship, winning, comic, won, critics, co, yard, scored

\textbf{Cluster 11}: son, captain, paul, thomas, father, john, george, manager, director, michael, david, edward, c., richard, james, charles, peter, william, scott, robert, martin, henry, president, a., lieutenant, mary, j., governor, mark, secretary, king, met, tom, wife, minister, member, sir, chris, prince, don, jack, lord, queen, al, van, mother, asked, saint, o, st, told, v., general, married, appointed, fort, de, born, la, und

\textbf{Cluster 12}: 1960, 1955, 1959, 1957, 1968, 1971, 1972, 1975, 1965, 1962, 1963, 1958, 1961, 1949, 1986, 1950, 1974, 1977, 1990, 1954, 1970, 1989, 1969, 1966, 1964, 1982, 1951, 1979, 1976, 1952, 1946, 1983, 1973, 1992, 1967, 1956, 1978, 1987, 1937, 1947, 1936, 1991, 1981, 1994, 1995, 1932, 1999, 1988, 1933, 1938, 1984, 1993, 1980, 1996, 1953, 1934, 1985, 1926, 1997, 1945, 2002, 1998, 1940, 1921, 1920, 2001, 1930, 2004, 1939, 2005, 1915, 1927, 2003, 1910, 1913, 2006, 1944, 2007, 1917, 1941, 1919, 1948, 1914, 1942, 2014, 1912, 2008, 2000, 2009, 1943, 1916, 2012, 2010, 2011, 2013, 2015, 1918, 2016, april, june, november, october, july, february, january, december, march, september, august, since, until, late

\section{Extended Vocabulary: 120 Clusters}
\label{app:full_120}
\noindent Configuration: $\gamma=0.2$, $k=120$.

\textbf{Cluster 0}: would, should, could, might, shall, will, must, can, did, not, does, able, unable, do, may

\textbf{Cluster 1}: here, appears, appear, there, ago, exactly

\textbf{Cluster 2}: men, people, women, children, players, man

\textbf{Cluster 3}: myself, else, them, you, everything, ourselves, yourself, him, someone, me, 'd, anything, forever, t, herself, we, something, somebody, characters, anyone, whoever, 're, really, everyone, nothing, just, ve, i, everybody, got, anybody, 'll, get, love, god, am, ll, hasn, ta

\textbf{Cluster 4}: mostly, solely, primarily, partly, mainly, largely, entirely, similarly, essentially, exclusively, specifically, extensively, partially, heavily, predominantly, historically, poorly, widely, completely, fully, strongly, closely, critically, commercially, locally, positively

\textbf{Cluster 5}: significantly, considerably, slightly, greatly, much, severely

\textbf{Cluster 6}: very, extremely, quite, too, equally, unusually, fairly, rather, somewhat, less, more, increasingly, truly, highly, relatively, pretty, politically, deeply

\textbf{Cluster 7}: be, been, were, being, was, is, had, previously, already, are, have, having, has, once, originally, now, recently, remains, currently, remained

\textbf{Cluster 8}: almost, either, well, virtually, nearly, far, species

\textbf{Cluster 9}: often, occasionally, sometimes, frequently, rarely, usually, always, normally, typically, merely, commonly, still, otherwise, regularly, generally, consistently, traditionally, easily, naturally, clearly, constantly, properly, longer, potentially, deliberately, physically, actively

\textbf{Cluster 10}: actually, indeed, also, therefore, nonetheless, nevertheless, consequently, probably, itself, apparently, likewise, himself, yet, themselves, certainly, it, others, even, thus, who, he, simply, initially, possibly, reportedly, they, perhaps, she, likely, never, literally, ever

\textbf{Cluster 11}: seriously, badly

\textbf{Cluster 12}: eventually, subsequently, again, then, finally, quickly, soon, simultaneously, instead, ultimately, immediately, effectively, briefly, suddenly, successfully, temporarily, rapidly, gradually, repeatedly, personally, slowly, permanently, officially, thereby, directly, formally, publicly, steadily, re, accidentally, unsuccessfully

\textbf{Cluster 13}: afterwards, afterward, meanwhile, whom, name

\textbf{Cluster 14}: young

\textbf{Cluster 15}: particularly, especially

\textbf{Cluster 16}: that, though, because, however, although, but, furthermore, if, where, whether, believed, said, and, why, wherein, when, thought, hence, saying, which, whenever, unless, nor, neither, additionally, how, stated, while, notably, so, both, what, whilst, felt

\textbf{Cluster 17}: know

\textbf{Cluster 18}: claimed

\textbf{Cluster 19}: became, become, becoming

\textbf{Cluster 20}: newly, formerly

\textbf{Cluster 21}: continued, planned

\textbf{Cluster 22}: following, during, prior

\textbf{Cluster 23}: used, created, added, made, making, given, make, provided, required, offered, developed, using, designed, based, proposed, received

\textbf{Cluster 24}: apart, involved, aside, reported, seen, found, upon, than, due, about, discovered, caused, support, responsible, amidst, without, amid, suffered

\textbf{Cluster 25}: result, suggested, addition, example, noted, revealed, notwithstanding, considered, despite, find, keep

\textbf{Cluster 26}: known

\textbf{Cluster 27}: performed, produced, worked, recorded, featured, wrote, appeared, composed, released, done, played, published, features, written, playing

\textbf{Cluster 28}: present, part

\textbf{Cluster 29}: arrived, leaving, moved, returned, captured, destroyed, left, formed, leave, built, placed, located, reached, damaged, killed, aboard

\textbf{Cluster 30}: going, go, come

\textbf{Cluster 31}: up, out, down, taken, place, back, removed, put, ran, away, held, off, take, took, came, taking, return, opened, carried, served, raised, turned, lost, running, established, changed, fell, went, takes, launched, passed, moving, working, extended, issued, sold, hit, close, closer

\textbf{Cluster 32}: sent, signed, selected, ordered, awarded

\textbf{Cluster 33}: to, wanted, allowed, decided, intended, helped, help, forced, agreed, failed

\textbf{Cluster 34}: former, fellow, old, leading

\textbf{Cluster 35}: with, including, included, as, alongside, for, called, behind, by, whereas, described, followed, include, compared, introduced, saw, replaced, see, except, unlike, like, led, besides, supported, gave, inspired, give, against, named, includes, brought, allow, e.g., directed, amongst, versus, joined, among, according, under, praised, defeated, opposite, ranked, vs.

\textbf{Cluster 36}: below, above

\textbf{Cluster 37}: from, of, through, into, on, in, over, towards, around, along, toward, within, across, onto, outside, at, between, near, via, beyond, entered, inside, nearby, throughout

\textbf{Cluster 38}: six, five, eight, four, three, seven, nine, twelve, two, ten, twenty, thirty, number, remaining

\textbf{Cluster 39}: many, several, other, few, numerous, various, some, these, all, multiple, most, those

\textbf{Cluster 40}: chief

\textbf{Cluster 41}: prime

\textbf{Cluster 42}: french, british, russian, german, spanish, italian, chinese, polish, american, english, australian, indian, canadian, japanese, african, european, soviet, local, independent, greek, roman, foreign, national, international, overseas, military, allied, imperial, professional, civil, federal, private, native, naval, christian, royal, senior, latin, armed

\textbf{Cluster 43}: grand

\textbf{Cluster 44}: 20th, 19th, 1st

\textbf{Cluster 45}: third, fourth, second, fifth, one, first, next, seventh, sixth, last, another, single, every, final, each, per, opening, top, pre

\textbf{Cluster 46}: only

\textbf{Cluster 47}: cut, long, short, open, double, hard, right, light, big, electric, super

\textbf{Cluster 48}: white, blue, red, black, green, yellow, brown, dead, dark, hot

\textbf{Cluster 49}: body

\textbf{Cluster 50}: large, larger, smaller, small, largest

\textbf{Cluster 51}: use, form, or, create, available, plus, provide

\textbf{Cluster 52}: high, level, higher, low, limited, increased, average, wide, overall, range, total, standard, highest, maximum, greater, heavy, lower, daily, free, deep

\textbf{Cluster 53}: previous, new, earlier, the, original, subsequent, current, future, recent, successful, early, official, regular, entire, traditional, annual

\textbf{Cluster 54}: female, male, unknown, living, powerful

\textbf{Cluster 55}: contemporary, modern, century, famous, prominent, older, popular, ancient, anti

\textbf{Cluster 56}: political, economic, social, cultural, legal, religious, historical, financial, public, medical

\textbf{Cluster 57}: musical

\textbf{Cluster 58}: associated, such, mixed

\textbf{Cluster 59}: similar, specific, particular, same, possible, any, related, important, significant, certain, key, different, individual, separate, potential, common, additional, no, minor, primary, main, critical, complex, active, strong, major, notable, initial, direct, special, commercial, positive, physical, natural, further, non, human, severe, nuclear

\textbf{Cluster 60}: this, whole, complete, full, a, an

\textbf{Cluster 61}: 's, own, whose, his, their, our, s, her, my, your, real, its, true, personal, great, greatest, secret

\textbf{Cluster 62}: good, kind, better, whatever, poor, difficult, enough, clear, bad, regardless, best, little

\textbf{Cluster 63}: worldwide

\textbf{Cluster 64}: 400, 300, 500, 800, 600, 250, 700, 200, 150, 000, 100, 120, thousand, hundred, million, billion

\textbf{Cluster 65}: 56, 44, 57, 46, 62, 43, 33, 39, 58, 66, 55, 42, 41, 35, 54, 53, 51, 32, 37, 48, 85, 45, 76, 65, 38, 70, 36, 34, 47, 80, 75, 52, 60, 40, 90, 95, 50, 110, 49, 64

\textbf{Cluster 66}: 22, 14, 16, 15, 17, 13, 26, 8, 27, 28, 12, 23, 9, 21, 19, 6, 24, 20, 7, 11, 25, 29, 10, 18, 5, 4, 30, 31, 3, 2, 1, 0

\textbf{Cluster 67}: feet, km, ft, mi, mph, m, h, c, miles, °, approximately, mm, roughly

\textbf{Cluster 68}: least

\textbf{Cluster 69}: estimated

\textbf{Cluster 70}: st., machine

\textbf{Cluster 71}: control

\textbf{Cluster 72}: base, front, ground, drive, side, field, fire, power, cross, forward, air, square, rear, water, aft

\textbf{Cluster 73}: training, offensive

\textbf{Cluster 74}: class

\textbf{Cluster 75}: set, forth, lead

\textbf{Cluster 76}: head, half

\textbf{Cluster 77}: run, pass, runs, point

\textbf{Cluster 78}: attack, battle, rest, fighting, operation

\textbf{Cluster 79}: line, bridge, station, road, building, park, airport, site, railway, center, street, hill, construction, highway, route

\textbf{Cluster 80}: system

\textbf{Cluster 81}: storm, hurricane, depression

\textbf{Cluster 82}: island, area, islands, region, areas, coast, bay, river, creek, ocean, sea, pacific, land, atlantic, damage, lake, ashore, offshore, mountain, population

\textbf{Cluster 83}: upper, middle

\textbf{Cluster 84}: south, east, north, northwest, west, northeast, eastern, southern, northern, western, northward, eastward, central

\textbf{Cluster 85}: corps, forces, brigade, force, army, battalion, infantry, regiment, squadron, division, troops, fleet, ships, command, ship, navy, aircraft, crew, guns, police

\textbf{Cluster 86}: ii, war, campaign, group, abroad, position, iii

\textbf{Cluster 87}: europe, africa, britain, canada, india, australia, france, germany, china, america, scotland, japan, kingdom, england, states, wales, territory, mexico, united, republic, country, elsewhere, empire, annually, uk, us

\textbf{Cluster 88}: council, society, government, congress, company, party, association, office, parliament, department, court, service, union, act, law, family

\textbf{Cluster 89}: city, town, hall, york, london, university, school, academy, college, house, chicago, county, angeles, district, boston, florida, virginia, california, texas, washington, state, michigan, bank, manchester, club, museum, church, home, oxford, los, studios

\textbf{Cluster 90}: san

\textbf{Cluster 91}: days, months, weeks, years, hours, beginning, month, day, time, year, start, ended, week, started, end, before, after, minutes, completed, announced, began, shortly, night, till, finished, died, summer, spent

\textbf{Cluster 92}: ahead

\textbf{Cluster 93}: mid

\textbf{Cluster 94}: en

\textbf{Cluster 95}: smith, jones, johnson, williams, jackson, hers, grant, members

\textbf{Cluster 96}: thereafter, later, twice

\textbf{Cluster 97}: won, winning, scored

\textbf{Cluster 98}: event, games, season, match, team, career, championship, win, cup, race, tour, league, points, awards, conference, award, victory, football, record, score, chart, yard

\textbf{Cluster 99}: player

\textbf{Cluster 100}: d, b, x

\textbf{Cluster 101}: co, post

\textbf{Cluster 102}: performance, film, show, release, game, series, book, version, story, song, music, album, studio, play, soundtrack, writing, songs, production, scene, band, episode, magazine, video, track, stage, style, recording, cast, design, action, title, sound, review, broadcast, character, screen, art, television, box, live, star, comic, radio, media, rock, entertainment

\textbf{Cluster 103}: work, events, history, works, success, life, development, records, list, project, times

\textbf{Cluster 104}: world

\textbf{Cluster 105}: critics

\textbf{Cluster 106}: u.s., alone, latter

\textbf{Cluster 107}: age

\textbf{Cluster 108}: order

\textbf{Cluster 109}: period, death

\textbf{Cluster 110}: onwards, respectively, role, case

\textbf{Cluster 111}: together, way

\textbf{Cluster 112}: shot, fight

\textbf{Cluster 113}: shows

\textbf{Cluster 114}: met, asked, told, appointed, born

\textbf{Cluster 115}: john, thomas, richard, george, james, william, david, paul, robert, edward, michael, peter, charles, henry, c., scott, j., a., martin, mary, mark, tom, chris, jack, don, sir, lord, van

\textbf{Cluster 116}: son, father, director, president, manager, captain, member, secretary, minister, governor, king, lieutenant, wife, prince, queen, saint, mother, al, st, o, general, married, de, v., fort, la, und

\textbf{Cluster 117}: october, june, july, december, april, november, january, september, february, august, march, until, since, late

\textbf{Cluster 118}: 1986, 1997, 1995, 1996, 1999, 1990, 1994, 1989, 2002, 1983, 1998, 1992, 2004, 1993, 1988, 1987, 2005, 2003, 1984, 2006, 2007, 2001, 1982, 1991, 1979, 1985, 1977, 1975, 2009, 1972, 1971, 1978, 1981, 2008, 1976, 1968, 2014, 1980, 2011, 1974, 2010, 1970, 2013, 1969, 2012, 1973, 1966, 1958, 1967, 2015, 2000, 2016

\textbf{Cluster 119}: 1937, 1936, 1946, 1949, 1938, 1932, 1947, 1955, 1933, 1934, 1940, 1957, 1926, 1951, 1921, 1952, 1950, 1961, 1956, 1939, 1915, 1913, 1959, 1963, 1930, 1919, 1945, 1920, 1914, 1965, 1960, 1912, 1917, 1944, 1916, 1953, 1910, 1941, 1962, 1964, 1954, 1942, 1943, 1927, 1948, 1918

\newpage

\section{Methods Details}
\label{app:methods}

\noindent This appendix gives the architecture specification, the composition of the training target, the base clustering settings, the metric definitions, and the transition-weight derivation underlying the Methods section of the main text. Nothing here is required to follow the results; it is provided so that every step of the pipeline can be reproduced.

\subsection{Residual Block Design Choices}
\label{app:methods_block}

Given an input vector $\mathbf{x} \in \mathbb{R}^{512}$, each \textit{ResidualBlock} computes
\begin{equation}
    \mathbf{x}' = \mathbf{x} + W_2\,\mathrm{GELU}\!\left(
        W_1\,\mathrm{LayerNorm}(\mathbf{x}) + \mathbf{b}_1
    \right) + \mathbf{b}_2,
    \label{eq:resblock}
\end{equation}
with learned weight matrices $W_1, W_2 \in \mathbb{R}^{512 \times 512}$ and bias vectors $\mathbf{b}_1, \mathbf{b}_2 \in \mathbb{R}^{512}$. This is the block drawn in the top-left panel of Figure~\mainFigArchi{} of the main text.

Pre-norm placement yields more stable gradients in deep networks than post-norm variants~\citep{xiong2020layer}, and GELU is preferred over ReLU for its smoother gradient signal~\citep{hendrycks2016gaussian}. No nonlinearity follows the residual addition, preserving unobstructed gradient flow.

The embedding table is $E \in \mathbb{R}^{20000 \times 512}$, mapping each token ID $t \in \{0, \ldots, 19{,}999\}$ to a learnable 512-dimensional dense vector, and the trunk of eight ResidualBlocks produces a position-wise hidden representation $H \in \mathbb{R}^{B \times 80 \times 512}$ for a batch of $B$ sequences of length 80. Each of the three heads consists of seven further ResidualBlocks followed by a linear projection $W_{\mathrm{out}} \in \mathbb{R}^{512 \times 20000}$ as its eighth layer and a softmax over the vocabulary axis, producing the normalised prediction
\begin{equation}
    \psi_\gamma = \mathrm{softmax}\!\left(W_{\mathrm{out}}\;\mathrm{BlockStack}_\gamma(H)\right)
    \;\in\; \mathbb{R}^{B \times 80 \times 20000}.
\end{equation}
The implementation retains the pre-softmax logits and evaluates both the loss and the softmax in log-space for numerical stability; $\psi_\gamma$ as used throughout denotes the normalised quantity, as in Figure~\mainFigArchi{} of the main text.
Because the blocks act independently at each sequence position, the batch and position axes carry no information across positions; they index independent predictions from the same token-level model.
The model has $56{,}283{,}744$ trainable parameters in total: $10.24$M in the embedding table, $4.21$M in the shared trunk, and $13.94$M in each of the three heads, of which $10.26$M is the output projection $W_{\mathrm{out}}$. The heads therefore hold roughly three quarters of the parameters, and the vocabulary-sized projections alone rather more than half.

The frozen shadow copy used for bootstrap targets is updated after every gradient step as
\begin{equation}
    \theta_{\mathrm{ema}} \leftarrow \alpha\,\theta_{\mathrm{ema}}
    + (1-\alpha)\,\theta_{\mathrm{live}},
\end{equation}
with $\alpha = 0.999$, matching the notation of Figure~\mainFigArchi{} of the main text.

\subsection{Composition of the \texorpdfstring{$\lambda$}{λ}-Return Target}
\label{app:methods_lambda}

Grouping the terms belonging to position $t$ leaves a geometric recursion with ratio
$\gamma\lambda$, namely $G_t = \left[\phi_{t+1} + \gamma(1-\lambda)\psi_{t+1}\right] +
\gamma\lambda\,G_{t+1}$, so that every step contributes one observed feature and one bootstrap
under the same damping. Unrolling it expresses the $\lambda$-return as an explicit weighting of
observed features against bootstrapped predictions,
\begin{equation}
    G_t = \sum_{k \ge 1} (\gamma\lambda)^{k-1}
          \left[\phi_{t+k} + \gamma\,(1-\lambda)\,\psi_{t+k}\right].
    \label{eq:lambda_unrolled}
\end{equation}
The two contributions can be compared by their total mass. The unscaled arrival \gls{sr} sums to
$(1-\gamma)^{-1}$, of which the observed features supply
$\sum_{k \ge 1}(\gamma\lambda)^{k-1} = (1-\gamma\lambda)^{-1}$ and the bootstrap terms the
remainder,
\begin{equation}
    \underbrace{\frac{1}{1-\gamma\lambda}}_{\text{observed}}
    \;+\;
    \underbrace{\frac{\gamma\,(1-\lambda)}{(1-\gamma\lambda)(1-\gamma)}}_{\text{bootstrapped}}
    \;=\; \frac{1}{1-\gamma}.
\end{equation}
The share of the target supplied by actually observed tokens is therefore
\begin{equation}
    \rho(\gamma,\lambda) = \frac{1-\gamma}{1-\gamma\lambda},
    \label{eq:rho}
\end{equation}
which at $\lambda = 0.9$ equals $0.976$, $0.909$ and $0.714$ for $\gamma = 0.2$, $0.5$ and $0.8$
respectively. The limiting cases are $\rho(\gamma,1) = 1$, a pure Monte Carlo target, and
$\rho(\gamma,0) = 1-\gamma$: the dial runs from $1-\gamma$ up to one rather than from zero,
because even the 1-step target contains the observed feature $\phi_{t+1}$. That floor is also the
immediate successor's own contribution at every $\lambda$, since the $k=1$ term of
Equation~\eqref{eq:lambda_unrolled} carries no factor of $\lambda$; $\rho$ rises above it only by
accumulating positions further ahead under the weights $(\gamma\lambda)^{k-1}$. It is in that sense
cumulative across the horizon rather than a statement about the next token: at $\gamma = 0.2$ the
observed share of $0.976$ resolves into $0.800$ contributed by the immediate successor and $0.176$
by all later positions together.

Scaling $\phi$ by $(1-\gamma)$ and normalising $\psi$ for the KL objective divides both
contributions equally,
\begin{equation}
    \frac{1-\gamma}{1-\gamma\lambda} \;+\; \frac{\gamma\,(1-\lambda)}{1-\gamma\lambda} \;=\; 1,
    \label{eq:rho_normalized}
\end{equation}
so $\rho$ is unchanged and can be read directly off the numerator; the denominator collapsing to
one is the statement in the Methods that the normalised target forms a valid distribution over the
vocabulary.

The two limits of $\lambda$ differ in what the type-level estimate converges to. At $\lambda = 0$
everything beyond the immediate successor is carried by $\psi$, a prediction already averaged over
the contexts in which the successor token occurs; the fixed point is then the \gls{sr} of the
first-order Markov chain over word types whose transition matrix is the empirical bigram matrix of
the corpus. At $\lambda = 1$ the target is the realised discounted future of the actual token
sequence, and the fixed point is the discounted occupancy that follows each word type, averaged
over every context in which that type occurs, which retains the long-range dependencies of the
corpus. The two coincide only if the sequence is Markovian in the word type. Because $\rho$
decreases in $\gamma$, the longest horizon relies on the bootstrap term roughly twelve times more
heavily than the shortest.

Equation~\eqref{eq:rho} measures the composition of the target, not the distance between the two
fixed points; it indicates where on the interpolation the training signal lies rather than bounding
the residual bias.

Padding contributes to the same target. Lines shorter than eighty tokens are padded to that length,
and no loss term is computed at a padded position. The last real token of a short line is still a
loss site, and its target falls largely on the padding that follows; the $\lambda$-return of earlier
positions accumulates the padded features too, and that share rises with the horizon as the
discounted window reaches further past the end of a short line: padding supplies $1.2\%$ of the
target mass at $\gamma = 0.2$, $1.7\%$ at $\gamma = 0.5$ and $2.4\%$ at $\gamma = 0.8$. The padding dimension is not removed from the objective
either, so the trained model places a small predicted mass there as well
(Appendix~\ref{app:unk_verify}).

\subsection{Base Clustering Settings}
\label{app:methods_baseclust}

Both base algorithms operate on the \gls{pca}-reduced \gls{sr} vectors under Euclidean distance, as
implemented in scikit-learn~\citep{pedregosa2011scikit}, the reduction retaining $99.99\%$ of the
variance as in the main text. Those vectors are the heads' logits;
Appendix~\ref{app:feature_scale} sets out what that scale is and what the alternatives to it give. $K$-Means uses ten random initialisations
per trial and keeps the solution with the lowest inertia. The Bayesian Gaussian mixtures use full
covariance matrices, three initialisations, and at most 300 expectation-maximisation iterations per
trial.

The mixture weights are governed by the library defaults: a Dirichlet process (stick-breaking) prior
with weight concentration $1/K_{\text{base}}$. Because that concentration decreases as the base
resolution grows, the prior favours sparse mixtures more strongly at the larger primes, and the
number of components receiving any assignment can fall below $K_{\text{base}}$. Only the resulting
hard assignments enter the co-association matrix, and the consensus hierarchy is cut at exactly $K$,
so this affects the diversity of the base partitions rather than the resolution of the reported
solutions.

\subsection{Clustering Metric Definitions}
\label{app:methods_metrics}

The three metrics are defined over a clustering \(\mathcal{C} = \{C_1, \dots, C_K\}\) and a reference partition \(\mathcal{L} = \{L_1, \dots, L_J\}\) over \(N\) tokens.

Purity is the fraction of tokens in a cluster belonging to its majority class,
\begin{equation}
  \mathrm{purity}(C_k) = \frac{1}{|C_k|} \max_{j} |C_k \cap L_j|.
\end{equation}
It is a per-cluster measure of homogeneity and is not comparable across resolutions, as it rises mechanically with $K$ and reaches one when $K = N$.

Two aggregates of it appear below. The paper reports the unweighted mean over clusters
throughout, written \emph{mean purity} or \emph{mean per-cluster purity},
\begin{equation}
  \overline{\mathrm{purity}}(\mathcal{C}) = \frac{1}{K} \sum_{k=1}^{K} \mathrm{purity}(C_k),
\end{equation}
which gives a cluster of two tokens the same weight as one of two hundred and is therefore
sensitive to fragmentation. Appendices~\ref{app:unk_gamma} and~\ref{app:feature_scale}
additionally compare against a size-weighted alternative, in which each cluster is weighted by its
share of the tokens and the factor $|C_k|$ cancels,
\begin{equation}
  \mathrm{purity}_{w}(\mathcal{C})
  = \sum_{k=1}^{K} \frac{|C_k|}{N}\,\mathrm{purity}(C_k)
  = \frac{1}{N} \sum_{k=1}^{K} \max_{j} |C_k \cap L_j|,
\end{equation}
leaving the fraction of all tokens that lie in the majority class of their own cluster. The two
coincide when clusters are of equal size and diverge as they are not.

\Gls{nmi}~\citep{strehl2002cluster} is the mutual information between the two partitions, normalised by their mean entropy,
\begin{equation}
  \mathrm{NMI}(\mathcal{C}, \mathcal{L})
  = \frac{2\,I(\mathcal{C};\mathcal{L})}{H(\mathcal{C}) + H(\mathcal{L})}.
\end{equation}
It penalises fragmentation and lies in \([0,1]\), but is not corrected for chance, so finer partitions can still achieve spuriously elevated scores.

\Gls{ari}~\citep{hubert1985comparing} corrects the Rand index for chance agreement under a
hypergeometric model,
\begin{equation}
  \mathrm{ARI}(\mathcal{C}, \mathcal{L})
  = \frac{\mathrm{RI} - \mathbb{E}[\mathrm{RI}]}
         {\max(\mathrm{RI}) - \mathbb{E}[\mathrm{RI}]},
\end{equation}
yielding a score in \([-1, 1]\) that is approximately resolution-independent, making it the most
reliable basis for comparison across values of $K$.

Writing $n_{kj} = |C_k \cap L_j|$ for the contingency table of the two partitions and
$a_k = |C_k|$, $b_j = |L_j|$ for its margins, the three terms expand to
\begin{equation}
  \mathrm{ARI}
  = \frac{\displaystyle\sum_{k,j}\binom{n_{kj}}{2}
          - \frac{\sum_k \binom{a_k}{2}\sum_j \binom{b_j}{2}}{\binom{N}{2}}}
         {\displaystyle\frac{1}{2}\left[\sum_k \binom{a_k}{2} + \sum_j \binom{b_j}{2}\right]
          - \frac{\sum_k \binom{a_k}{2}\sum_j \binom{b_j}{2}}{\binom{N}{2}}}.
  \label{eq:ari_full}
\end{equation}
The subtracted term is the number of co-assigned pairs expected when the two partitions are drawn
independently with their margins held fixed. This is the whole of the chance correction, and it is
what \gls{nmi} lacks: splitting a cluster raises $\mathbb{E}[\mathrm{RI}]$ along with
$\mathrm{RI}$, so a partition unrelated to the reference scores near zero at any $K$, whereas its
\gls{nmi} rises with $K$. The correction fixes the baseline, not the value: both metrics still move
with resolution on the partitions reported here, and \gls{ari} falls the more steeply of the two as
the solution fragments away from the reference. All three metrics are computed
with the scikit-learn implementations~\citep{pedregosa2011scikit} at their default settings, so the
\gls{nmi} normalisation is the arithmetic mean of the two entropies given above.

\subsection{Derivation of the Inter-Cluster Transition Weights}
\label{app:methods_transitions}

For a source token $u$ the head at discount factor $\gamma$ yields the predicted successor distribution $\psi_\gamma(u)$ over the vocabulary, with successor representation $\mathrm{SR}_\gamma(u)=\psi_\gamma(u)\,(1\!-\!\gamma)^{-1}$. Restricting targets to the token set $\mathcal{T}$ entering the clustering, the pooled predicted mass from cluster $C_i$ towards cluster $C_j$ is
\begin{equation}
    A_{ij}=\frac{1}{|C_i|}\sum_{u\in C_i}\sum_{v\in C_j}\mathrm{SR}_\gamma(u)_v.
\end{equation}
The source cluster's size $|C_i|$ and the discount rescaling $(1\!-\!\gamma)^{-1}$ are constant within a row and therefore cancel when the row is normalised,
\begin{equation}
M_{ij}=\frac{A_{ij}}{\sum_{j'}A_{ij'}}
      =\frac{\tfrac{1}{|C_i|}(1\!-\!\gamma)^{-1}\sum_{u\in C_i}\sum_{v\in C_j}\psi_\gamma(u)_v}
            {\tfrac{1}{|C_i|}(1\!-\!\gamma)^{-1}\sum_{u\in C_i}\sum_{v\in \mathcal{T}}\psi_\gamma(u)_v}
      =\frac{\sum_{u\in C_i}\sum_{v\in C_j}\psi_\gamma(u)_v}
            {\sum_{u\in C_i}\sum_{v\in \mathcal{T}}\psi_\gamma(u)_v},
\end{equation}
so that $M_{ij}$ is the share of the pooled successor mass of cluster $i$ that falls on cluster $j$, and rows sum to one over all target clusters in $\mathcal{T}$. All reported weights come from the $\gamma = 0.2$ head.

\section{Robustness of the Reported Clustering}
\label{app:unk}

This appendix reports how far the clustering results depend on four sources of variability: which
dimensions of the predicted successor distribution enter the clustering, the random seed of the
consensus procedure, the training run itself, and the scale on which the read-out is clustered. The
first, second and fourth are choices the analysis leaves open; the third is the variability the
reported model is one draw from, assessed against two further models trained under different seeds.
The dependence on the predictive horizon is checked against the first and the fourth in turn.

\subsection{Reserved Vocabulary Dimensions}
\label{app:unk_verify}

The released vocabulary reserves seventeen entries of the form \texttt{<unk\_NOUN>},
\texttt{<unk\_VERB>}, \texttt{<unk\_PROPN>} and so on, at token IDs 3--19. These are residues of an
earlier preprocessing design in which out-of-vocabulary words were to be replaced by
part-of-speech-specific unknown tags rather than by a single generic one. A case mismatch between the
preprocessed corpus and the tokeniser's vocabulary lookup meant that they never matched, so the model
was trained with the single generic unknown token described in the main text, and these seventeen
dimensions receive no probability mass.

They nonetheless remain as dimensions of the predicted successor distribution on which the clustering
operates. Counting them together with the padding symbol, the end-of-sequence symbol and the generic
\texttt{<unk>} token, twenty of the 20{,}000 output dimensions are reserved in this way. Only two carry
appreciable predicted mass at $\gamma=0.2$: \texttt{<unk>} with $3.49\%$, and padding with $0.05\%$,
the latter reflecting the padded positions that enter the discounted target at the end of a short line.
The remaining eighteen, the seventeen unused entries and the end-of-sequence symbol, carry none.
Because out-of-vocabulary words are not uniformly distributed across categories, the reserved
dimensions are not entirely uninformative about category: part of speech is recoverable from the
\texttt{<unk>} dimension alone at $0.48$, and from all twenty reserved dimensions jointly at $0.64$,
against a majority-class rate of $0.333$. Twenty ordinary word dimensions drawn at random support
$0.79$ on average (range $0.74$--$0.83$ over twenty draws), so the reserved dimensions carry less
category information than an equal number of ordinary ones rather than none at all. The next section
establishes that removing them leaves the reported agreement with part-of-speech labels essentially
unchanged.

\subsection{Effect of Removing the Unknown Dimensions}
\label{app:unk_ablation}

The full clustering pipeline was re-run with the unknown dimensions removed, under four dimension conditions: all dimensions retained, \texttt{<unk>} removed, the seventeen unused dimensions removed,
and all twenty removed. Preprocessing and clustering are otherwise identical to the main analysis.
Each condition is evaluated at fourteen cluster counts between 3 and 100 for the
noun--verb--adjective (NVA) subset, and at sixteen between 13 and 200 for the extended vocabulary,
the latter starting at the number of reference categories because any coarser solution must merge
distinct ones. Both are subsets of the resolution scans of the main analysis, which re-cut the
consensus hierarchy at every integer $K$ over these ranges. The coarser sampling is chosen to
compare the four conditions at equal cost rather than to reproduce those scans, and every statement
below therefore concerns agreement between conditions at matched resolutions, not the shape of the
curve itself.
Table~\ref{tab:unk_ablation} reports the two headline resolutions. Its baseline column re-extracts and
re-clusters the same checkpoint under the same settings and the same random seed, and reproduces the
values reported in the main text, so the columns can be read against the main analysis as well as
against one another.

\begin{table}[htbp]
\centering
\small
\caption{Agreement with part-of-speech labels under removal of the unknown dimensions, $\gamma=0.2$.
Values are mean per-cluster purity, normalised mutual information and adjusted Rand index against the
reference labels.}
\label{tab:unk_ablation}
\begin{tabular}{llcccc}
\toprule
& & \textbf{all dims} & $-$\texttt{<unk>} & $-$\textbf{unused} & $-$\textbf{all 20} \\
\midrule
\multicolumn{6}{l}{\textit{NVA subset, $K=3$, consensus KMeans}} \\
& purity & 0.8816 & 0.8605 & 0.8685 & 0.8846 \\
& NMI    & 0.5954 & 0.5520 & 0.5740 & 0.6040 \\
& ARI    & 0.6714 & 0.6254 & 0.6417 & 0.6801 \\
\midrule
\multicolumn{6}{l}{\textit{NVA subset, $K=3$, consensus BGM}} \\
& purity & 0.8829 & 0.8896 & 0.8881 & 0.8863 \\
& NMI    & 0.6022 & 0.6199 & 0.6167 & 0.6108 \\
& ARI    & 0.6722 & 0.6885 & 0.6844 & 0.6802 \\
\midrule
\multicolumn{6}{l}{\textit{Extended vocabulary, $K=13$, consensus KMeans}} \\
& purity & 0.6155 & 0.6103 & 0.6080 & 0.6160 \\
& NMI    & 0.5077 & 0.5039 & 0.5022 & 0.5057 \\
& ARI    & 0.4107 & 0.4076 & 0.4049 & 0.4070 \\
\midrule
\multicolumn{6}{l}{\textit{Extended vocabulary, $K=13$, consensus BGM}} \\
& purity & 0.6036 & 0.6058 & 0.6156 & 0.6264 \\
& NMI    & 0.5157 & 0.5144 & 0.5100 & 0.5111 \\
& ARI    & 0.4209 & 0.4195 & 0.4173 & 0.4141 \\
\bottomrule
\end{tabular}
\end{table}

Across every cluster count evaluated, and across both consensus algorithms, the largest change in
normalised mutual information produced by removing the unknown dimensions is $0.055$
(NVA subset, $\gamma=0.5$, \glspl{bgm}) and below $0.044$ in all eleven other
algorithm-by-vocabulary-by-head combinations. Removal moves agreement upward in $292$ of the $540$
comparisons and downward in the remaining $248$, so the reported geometry is not an artefact inflated
by these dimensions.

\subsection{Stability of the Reported Partition}
\label{app:unk_seed}

The reported solution is one draw from a procedure with several sources of variability. This appendix
separates three of them at the four resolutions whose cluster listings this Supplementary Material
provides: removing reserved vocabulary dimensions from the clustering features, re-running the
consensus aggregation under a different random seed, and re-training the network from scratch under a
different seed. For the last, two further models were trained, identical to the reported one in every
setting of Table~\mainTabHyper{} of the main text but the seed. Their final training losses are $3.7104$ and
$3.7111$ against the reported model's $3.7104$, and their validation losses $3.7455$ and $3.7460$
against $3.7453$, so the optimisation itself is highly reproducible; what follows concerns the
geometry it produces.
Table~\ref{tab:seed_stability} summarises the stability of the reported partitions under these three perturbations.

\begin{widetable}[htbp]
\centering
\caption{Stability of the consensus partition under three perturbations, consensus $K$-Means at
$\gamma = 0.2$. The left block gives the adjusted Rand index against the reported partition, the
right block the adjusted Rand index against the \gls{pos} labels. Individual values are listed rather
than summarised: with three dimension variants, three consensus seeds and two training runs, a range or
a median would imply more precision than the counts support.}
\label{tab:seed_stability}
\small
\begin{fitwidth}
\begin{tabular}{lcccccccc}
\toprule
 & \multicolumn{4}{c}{\gls{ari} vs.\ reported partition} & \multicolumn{4}{c}{\gls{ari} vs.\ \gls{pos}} \\
\cmidrule(lr){2-5}\cmidrule(lr){6-9}
 & \multicolumn{2}{c}{NVA} & \multicolumn{2}{c}{Ext.} & \multicolumn{2}{c}{NVA} & \multicolumn{2}{c}{Ext.} \\
\cmidrule(lr){2-3}\cmidrule(lr){4-5}\cmidrule(lr){6-7}\cmidrule(lr){8-9}
Perturbation & $K{=}3$ & $K{=}30$ & $K{=}13$ & $K{=}120$ & $K{=}3$ & $K{=}30$ & $K{=}13$ & $K{=}120$ \\
\midrule
Reported solution              & $1.000$ & $1.000$ & $1.000$ & $1.000$ & $0.671$ & $0.145$ & $0.411$ & $0.137$ \\
\midrule
Features, no \texttt{<unk>}    & $0.811$ & $0.861$ & $0.937$ & $0.950$ & $0.625$ & $0.140$ & $0.408$ & $0.141$ \\
Features, no \texttt{<unk\_POS>} & $0.829$ & $0.824$ & $0.932$ & $0.889$ & $0.642$ & $0.128$ & $0.405$ & $0.141$ \\
Features, no reserved          & $0.980$ & $0.761$ & $0.918$ & $0.892$ & $0.680$ & $0.137$ & $0.407$ & $0.140$ \\
\midrule
Consensus seed 0               & $0.945$ & $0.744$ & $0.902$ & $0.866$ & $0.675$ & $0.146$ & $0.415$ & $0.133$ \\
Consensus seed 7               & $0.927$ & $0.772$ & $0.894$ & $0.840$ & $0.659$ & $0.148$ & $0.424$ & $0.139$ \\
Consensus seed 123             & $0.889$ & $0.755$ & $0.867$ & $0.871$ & $0.676$ & $0.160$ & $0.397$ & $0.143$ \\
\midrule
Retrained, seed 2              & $0.908$ & $0.698$ & $0.893$ & $0.803$ & $0.652$ & $0.152$ & $0.422$ & $0.139$ \\
Retrained, seed 3              & $0.926$ & $0.670$ & $0.862$ & $0.733$ & $0.680$ & $0.147$ & $0.439$ & $0.152$ \\
\bottomrule
\end{tabular}
\end{fitwidth}
\end{widetable}

Two quantities separate in Table~\ref{tab:seed_stability}. Agreement with the reported partition
falls under every perturbation, to
between $0.67$ and $0.98$, so no perturbation leaves the memberships intact. Agreement with the
\gls{pos} labels barely moves: it stays within a few hundredths of the reported value in every cell,
and exceeds it in several, including both replicates on the extended vocabulary at $K = 13$. What the
partition is measured to achieve is stable even where the partition itself is not.

The perturbations differ with resolution rather than in overall size. At $K = 3$ and $K = 13$, the
resolutions at which the main text reports solutions, re-training displaces the partition no more
than re-seeding the aggregation does, at an \gls{ari} of $0.908$ and $0.926$ against a re-seeding
range of $0.889$ to $0.945$, and $0.893$ and $0.862$ against $0.867$ to $0.902$. At $K = 30$ and $K = 120$ it displaces
it further, and the two no longer overlap. The coarse solutions the main text presents as findings
are thus reproducible across independent training runs to the same degree that they are reproducible
across re-runs of the clustering. The finer solutions, presented as exploratory, reproduce less
closely, at $0.67$ to $0.80$: well above chance, but below the $0.74$ to $0.87$ that re-seeding the
aggregation alone produces at those resolutions.

The aggregate quality measures are correspondingly more stable than the partition. Across the four
consensus seeds, mean purity at $K = 3$ in the NVA subset ranges over $0.8798$--$0.8851$ and
normalised mutual information over $0.5927$--$0.6048$; at $K = 13$ in the extended vocabulary, purity
ranges over $0.6155$--$0.6391$ and normalised mutual information over $0.5077$--$0.5231$. Across the
two retrained replicates the same quantities stay within those ranges.

Cluster-level descriptions in the main text should therefore be read as characterising one
representative solution. The groupings that recur across subsets and resolutions, such as the
temporal cluster combining duration nouns with aspectual verbs, carry correspondingly more weight
than any single cluster's exact membership.

\subsection{Robustness of the Horizon Effect to the Reserved Dimensions}
\label{app:unk_gamma}

The decline of \gls{pos} agreement with increasing $\gamma$ is the claim most exposed to the choice of
dimensions, since it is the only one compared across heads. Table~\ref{tab:unk_gamma} therefore repeats
it under all four dimension conditions, for both consensus algorithms, both vocabularies and all three
metrics. Each entry is one consensus run cut at every evaluated cluster count and averaged over those
cuts, not an average over repetitions. Cluster counts below the number of reference categories are not
evaluated for the extended vocabulary, since any such solution is forced to merge distinct categories.

\begin{widetable}[htbp]
\centering
\small
\setlength{\tabcolsep}{4pt}
\caption{Mean purity, \gls{nmi} and \gls{ari}, averaged across evaluated cluster counts, by head and
dimension condition, for both consensus algorithms. The ordering $\gamma=0.2 > 0.5 > 0.8$ holds in every
row for every metric.}
\label{tab:unk_gamma}
\begin{fitwidth}
\begin{tabular}{llccccccccc}
\toprule
& & \multicolumn{3}{c}{Mean purity} & \multicolumn{3}{c}{\gls{nmi}} & \multicolumn{3}{c}{\gls{ari}} \\
\cmidrule(lr){3-5} \cmidrule(lr){6-8} \cmidrule(lr){9-11}
& $\gamma$ & $0.2$ & $0.5$ & $0.8$ & $0.2$ & $0.5$ & $0.8$ & $0.2$ & $0.5$ & $0.8$ \\
\midrule
\multicolumn{11}{l}{\textit{Consensus KMeans}} \\
\midrule
\multirow{4}{*}{\rotatebox{90}{NVA}}
& all dimensions       & 0.8934 & 0.8044 & 0.6883 & 0.3910 & 0.2615 & 0.1442 & 0.2107 & 0.1088 & 0.0434 \\
& $-$\texttt{<unk>}    & 0.8933 & 0.8003 & 0.6906 & 0.3855 & 0.2562 & 0.1446 & 0.2041 & 0.1026 & 0.0441 \\
& $-$unused            & 0.8879 & 0.7912 & 0.6843 & 0.3919 & 0.2544 & 0.1439 & 0.2088 & 0.1008 & 0.0468 \\
& $-$all 20            & 0.8903 & 0.8009 & 0.6860 & 0.3926 & 0.2603 & 0.1395 & 0.2130 & 0.1044 & 0.0427 \\
\midrule
\multirow{4}{*}{\rotatebox{90}{Extended}}
& all dimensions       & 0.7525 & 0.7423 & 0.6656 & 0.4959 & 0.4439 & 0.3644 & 0.2052 & 0.1675 & 0.1146 \\
& $-$\texttt{<unk>}    & 0.7514 & 0.7398 & 0.6661 & 0.4953 & 0.4452 & 0.3654 & 0.2031 & 0.1684 & 0.1176 \\
& $-$unused            & 0.7692 & 0.7394 & 0.6679 & 0.4936 & 0.4408 & 0.3659 & 0.1992 & 0.1652 & 0.1176 \\
& $-$all 20            & 0.7624 & 0.7439 & 0.6617 & 0.4936 & 0.4433 & 0.3647 & 0.2010 & 0.1663 & 0.1176 \\
\midrule
\multicolumn{11}{l}{\textit{Consensus \glspl{bgm}}} \\
\midrule
\multirow{4}{*}{\rotatebox{90}{NVA}}
& all dimensions       & 0.8869 & 0.7906 & 0.6888 & 0.3913 & 0.2528 & 0.1425 & 0.2166 & 0.1062 & 0.0420 \\
& $-$\texttt{<unk>}    & 0.8898 & 0.7838 & 0.6843 & 0.3966 & 0.2539 & 0.1415 & 0.2212 & 0.1077 & 0.0415 \\
& $-$unused            & 0.8774 & 0.7926 & 0.6750 & 0.3962 & 0.2535 & 0.1369 & 0.2250 & 0.1074 & 0.0394 \\
& $-$all 20            & 0.8752 & 0.8025 & 0.6693 & 0.3914 & 0.2639 & 0.1394 & 0.2197 & 0.1139 & 0.0412 \\
\midrule
\multirow{4}{*}{\rotatebox{90}{Extended}}
& all dimensions       & 0.7511 & 0.7259 & 0.6516 & 0.4951 & 0.4391 & 0.3618 & 0.2130 & 0.1679 & 0.1188 \\
& $-$\texttt{<unk>}    & 0.7448 & 0.7323 & 0.6555 & 0.4969 & 0.4395 & 0.3641 & 0.2108 & 0.1685 & 0.1214 \\
& $-$unused            & 0.7474 & 0.7323 & 0.6549 & 0.4971 & 0.4434 & 0.3637 & 0.2136 & 0.1747 & 0.1230 \\
& $-$all 20            & 0.7500 & 0.7280 & 0.6554 & 0.4959 & 0.4425 & 0.3681 & 0.2108 & 0.1693 & 0.1229 \\
\bottomrule
\end{tabular}
\end{fitwidth}
\end{widetable}

The four dimension conditions are interchangeable. Within any one head, the largest difference between
two conditions anywhere in the table is $0.019$ for mean purity, $0.011$ for \gls{nmi} and $0.009$ for
\gls{ari}. Measured on the same averages, re-seeding the consensus aggregation moves mean purity by up
to $0.019$ and \gls{ari} by up to $0.016$, so for those two metrics the dimension choice matters about
as much as the seed does; for \gls{nmi}, whose seed-to-seed range is $0.005$, it matters about twice
as much.
Either way the perturbation is an order of magnitude below the effect it might disturb: across these
same averages the drop from $\gamma=0.2$ to $\gamma=0.8$ is $0.085$--$0.206$ for mean purity,
$0.128$--$0.259$ for \gls{nmi} and $0.082$--$0.186$ for \gls{ari}. The ordering
$\gamma = 0.2 > 0.5 > 0.8$ holds in all sixteen rows of the table for all three metrics, that is in
all forty-eight orderings, without exception.

Read cut by cut rather than on the average, the ordering holds at every one of the $240$ evaluated
cluster counts for \gls{nmi} and for \gls{ari}. Mean purity holds at $215$ of the $240$, and all
twenty-five exceptions lie in the extended vocabulary, in two bands at $K=30$--$70$ and
$K \approx 140$--$180$; the resolutions reported in the main text, $K=13$ and $K=120$, are not among
them. Every exception is an inversion between the two closest heads, $\gamma=0.2$ and $\gamma=0.5$,
by a median margin of $0.009$ and never more than $0.055$; the step from $\gamma=0.5$ to $\gamma=0.8$
never inverts. Nor are the inversions produced by the ablation---five of them occur under the
all-dimensions baseline, while one ablated condition under the same algorithm has none---so they are a
property of mean per-cluster purity near a tie at those resolutions rather than a consequence of
removing dimensions. Mean per-cluster purity weights every cluster equally regardless of size, and
weighting by size removes them: across the full resolution scan behind Figures~\mainFigAriNVA{} and~\mainFigAriFull{} of the main
text the ordering then holds at all $188$ cluster counts at or above $K=13$ under both consensus
algorithms, against $118$ of $188$ unweighted for consensus KMeans and $176$ of $188$ for the
consensus \glspl{bgm}.

\subsection{Feature Scale: Logits, Log-Ratios and the Successor Distribution}
\label{app:feature_scale}

The vectors clustered throughout are the heads' logits, whereas the successor representation is
$\psi_\gamma$ rescaled by $(1-\gamma)^{-1}$. The two are one object on two scales.
Writing $z$ for a head's pre-softmax output at a source token and $c = \log \sum_v \exp z_v$ for its
log-normaliser,
\begin{equation}
  z \;=\; \log \psi_\gamma + c \;=\; \log \mathrm{SR}_\gamma + \log(1-\gamma) + c ,
  \label{eq:scale_identity}
\end{equation}
so the analysed vectors are the log successor representation displaced by one constant per word.
This appendix asks what that displacement, and the logarithm itself, are worth. Four feature sets are
pushed through the pipeline of Appendix~\ref{app:methods_baseclust} unchanged, at every horizon and in
both vocabularies:

\begin{itemize}
  \item $z$, the reported read-out.
  \item $\mathrm{clr}(\psi_\gamma)_v = \log \psi_\gamma(v) - \frac{1}{|V|}\sum_u \log \psi_\gamma(u)$,
    the centred log-ratio coordinates in which compositional data carry a Euclidean
    geometry~\citep{aitchison1982statistical}. Two identities make this the reference condition. The
    per-word constant cancels, $\mathrm{clr}(\psi_\gamma) = z - \frac{1}{|V|}\sum_u z_u$, so these
    coordinates are a function of the successor distribution alone; and $\mathrm{clr}$ is invariant to
    a positive rescaling, so $\mathrm{clr}(\mathrm{SR}_\gamma) = \mathrm{clr}(\psi_\gamma)$ exactly.
  \item $\log \psi_\gamma$, the log successor representation itself, up to the global $\log(1-\gamma)$.
  \item $\psi_\gamma$, the normalised successor distribution. The rescaling to $\mathrm{SR}_\gamma$
    multiplies every vector by one
    positive scalar and so cannot change the geometry, and a confirmatory run bears this out below.
    The two are not distinguished here.
\end{itemize}

Because $\psi_\gamma$ is a softmax it is strictly positive, so the logarithms need no smoothing floor,
unlike the count-based vectors of Appendix~\ref{app:baselines}.

Table~\ref{tab:scale_variance} measures the property the \gls{pca} and the Euclidean clustering are
sensitive to: how the variance across words is distributed over the vocabulary entries. On $z$,
$\mathrm{clr}(\psi_\gamma)$ and $\log \psi_\gamma$ it is spread almost evenly, the ten most variable
of the $20{,}000$ entries carrying about $0.1\%$ of the total and the hundred most variable about
$1\%$. On $\psi_\gamma$ the same information is concentrated onto a handful of entries, ten carrying
roughly two-thirds of the variance and a hundred carrying $94$ to $96\%$: words differ mainly in the
probability they place on a few frequent successors. The softmax is monotone within a row and leaves
the ordering of a word's successors unchanged; what it changes is which coordinates a Euclidean metric
can see. The reduction does not repair this. Reaching the $99.99\%$ variance threshold takes $508$ to
$510$ components on the log-scale features of the extended vocabulary but $1{,}040$ to $1{,}097$ on
$\psi_\gamma$: a concentrated head over a long flat tail forces \gls{pca} to keep twice as many
directions. The raw scale is thus reduced to roughly twice the rank of the others, and still fails.

\begin{widetable}[htbp]
\centering
\small
\caption{Concentration of variance across the $20{,}000$ vocabulary entries, by feature scale, head
and vocabulary. \emph{Top 10} and \emph{Top 100} give the share of the total variance carried by the
ten and the hundred most variable vocabulary entries. \emph{PCA comp.} is the number of principal
components the reduction retains at the $99.99\%$ variance threshold used throughout.}
\label{tab:scale_variance}
\begin{fitwidth}
\begin{tabular}{llcccccc}
\toprule
& & \multicolumn{3}{c}{NVA} & \multicolumn{3}{c}{Extended} \\
\cmidrule(lr){3-5}\cmidrule(lr){6-8}
Scale & $\gamma$ & Top 10 & Top 100 & PCA comp. & Top 10 & Top 100 & PCA comp. \\
\midrule
\multirow{3}{*}{$z$ (reported)}
 & $0.2$ & $0.0010$ & $0.0090$ & $488$ & $0.0013$ & $0.0103$ & $509$ \\
 & $0.5$ & $0.0011$ & $0.0094$ & $490$ & $0.0014$ & $0.0112$ & $509$ \\
 & $0.8$ & $0.0012$ & $0.0104$ & $487$ & $0.0016$ & $0.0124$ & $508$ \\
\midrule
\multirow{3}{*}{$\mathrm{clr}(\psi_\gamma)$}
 & $0.2$ & $0.0010$ & $0.0090$ & $488$ & $0.0014$ & $0.0109$ & $509$ \\
 & $0.5$ & $0.0011$ & $0.0094$ & $490$ & $0.0015$ & $0.0118$ & $509$ \\
 & $0.8$ & $0.0012$ & $0.0106$ & $487$ & $0.0017$ & $0.0132$ & $508$ \\
\midrule
\multirow{3}{*}{$\log \psi_\gamma$}
 & $0.2$ & $0.0010$ & $0.0091$ & $487$ & $0.0011$ & $0.0094$ & $510$ \\
 & $0.5$ & $0.0010$ & $0.0093$ & $490$ & $0.0012$ & $0.0101$ & $510$ \\
 & $0.8$ & $0.0011$ & $0.0102$ & $487$ & $0.0014$ & $0.0116$ & $509$ \\
\midrule
\multirow{3}{*}{$\psi_\gamma$}
 & $0.2$ & $0.6687$ & $0.9597$ & $518$ & $0.6610$ & $0.9435$ & $1{,}040$ \\
 & $0.5$ & $0.6471$ & $0.9546$ & $533$ & $0.6445$ & $0.9400$ & $1{,}097$ \\
 & $0.8$ & $0.5976$ & $0.9500$ & $537$ & $0.6120$ & $0.9383$ & $1{,}080$ \\
\bottomrule
\end{tabular}
\end{fitwidth}
\end{widetable}

Table~\ref{tab:feature_scale} reports the four showcase resolutions at $\gamma = 0.2$. Two results
stand out, and they answer different questions.

\begin{widetable}[htbp]
\centering
\small
\setlength{\tabcolsep}{4pt}
\caption{\gls{ari}, mean per-cluster purity, and agreement with the reported partition, by feature
scale at $\gamma = 0.2$, $K$-Means consensus, at the four showcase resolutions. Agreement is the
\gls{ari} between the solution and the deposited one at the same cluster count.}
\label{tab:feature_scale}
\begin{fitwidth}
\begin{tabular}{lcccccccccccc}
\toprule
& \multicolumn{3}{c}{NVA $K=3$} & \multicolumn{3}{c}{NVA $K=30$}
& \multicolumn{3}{c}{Ext.\ $K=13$} & \multicolumn{3}{c}{Ext.\ $K=120$} \\
\cmidrule(lr){2-4}\cmidrule(lr){5-7}\cmidrule(lr){8-10}\cmidrule(lr){11-13}
Scale & \gls{ari} & Pur. & Agr. & \gls{ari} & Pur. & Agr. & \gls{ari} & Pur. & Agr.
      & \gls{ari} & Pur. & Agr. \\
\midrule
$z$ (reported)              & $0.671$ & $0.882$ & $1.000$ & $0.145$ & $0.872$ & $1.000$
                            & $0.411$ & $0.615$ & $1.000$ & $0.137$ & $0.806$ & $1.000$ \\
$\mathrm{clr}(\psi_\gamma)$ & $0.722$ & $0.901$ & $0.894$ & $0.155$ & $0.882$ & $0.695$
                            & $0.413$ & $0.623$ & $0.840$ & $0.146$ & $0.814$ & $0.795$ \\
$\log \psi_\gamma$          & $0.565$ & $0.860$ & $0.702$ & $0.164$ & $0.860$ & $0.431$
                            & $0.330$ & $0.662$ & $0.623$ & $0.154$ & $0.803$ & $0.594$ \\
$\psi_\gamma$               & $0.008$ & $0.629$ & $0.011$ & $0.327$ & $0.880$ & $0.125$
                            & $0.165$ & $0.633$ & $0.216$ & $0.208$ & $0.825$ & $0.216$ \\
\bottomrule
\end{tabular}
\end{fitwidth}
\end{widetable}

\paragraph{The reported geometry is the geometry of the successor distribution.}
Clustering $\mathrm{clr}(\psi_\gamma)$ --- a function of the successor distribution alone, identical
for $\mathrm{SR}_\gamma$, and independent of the constant $c$ --- matches or slightly exceeds the
reported solution at all four showcase resolutions on both \gls{ari} and mean purity, at $0.722$
against $0.671$ on the NVA subset at $K=3$ and $0.413$ against $0.411$ on the extended
vocabulary at $K=13$. It also reproduces the reported partitions closely, though not uniformly
within the band that re-seeding the consensus aggregation alone produces
(Appendix~\ref{app:unk_seed}). On the NVA subset at $K = 3$ it agrees at $0.894$, inside the
re-seeding range of $0.889$ to $0.945$, and on the extended vocabulary at $K = 120$ at $0.795$,
inside the $0.74$ to $0.87$ that re-seeding produces at the fine resolutions. At the other two it
falls a little below: $0.840$ at $K = 13$ against a re-seeding range of $0.867$ to $0.902$, and
$0.695$ at $K = 30$ against $0.74$ to $0.87$. Across the whole resolution scan the median agreement
is $0.79$ on the NVA subset and $0.80$ on the extended vocabulary. The reported analysis therefore
does not depend on reading a quantity outside the successor distribution: in the coordinates that
carry a Euclidean geometry for distributions, the same categories emerge at the same strength in
nearly the same partition.

\paragraph{The logarithm is what makes the geometry legible, not the network.}
On the untransformed $\psi_\gamma$ the three classes are not recovered at all, at an \gls{ari} of
$0.008$ where the reported solution reaches $0.671$, and the resulting partition is unrelated to it
rather than a degraded version of it, agreeing at $0.011$. Its apparent purity of $0.629$ there is an
artefact of equal weighting: the solution is one cluster of $532$ tokens at a purity of $0.357$
beside two of $35$ and $33$ at $0.771$ and $0.758$, so the unweighted mean is carried by the two
small clusters. Weighted by size it is $0.403$, against $0.880$ for the reported read-out, on a
subset whose three categories are equally sized.
This reproduces on the trained model the
effect Appendix~\ref{app:baselines} reports for the analytic \gls{sr} computed from corpus counts,
which scores $0.017$ untransformed against $0.639$ under a logarithm. Occupancies compose
multiplicatively along a path, so a Euclidean metric on them has no additive structure to exploit; the
logarithm is the transformation under which discounted occupancy ratios become
distances~\citep{myers2024learning}, and the log-occupancy of a successor relative to its base rate is
the quantity that connects predictive maps to the pointwise mutual information underlying distributional
word embeddings~\citep{haga2025unified,levy2014neural}. The failure at $K=3$ is a property of the distance, not of the representation. Rescaling
$\psi_\gamma$ to $\mathrm{SR}_\gamma$ changes none of it, as it cannot: the confirmatory run at
$\gamma = 0.2$ returns the identical partition at $K = 3$, and at $K = 30$ one agreeing at $0.979$,
the small residual arising from a variance threshold that retains $519$ components rather than
$518$.

\paragraph{The horizon effect survives the scale change.}
The decline of agreement with increasing $\gamma$ is the one claim compared across heads.
Table~\ref{tab:scale_horizon} counts the resolutions at which the ordering
$\gamma = 0.2 > 0.5 > 0.8$ holds on each scale, cut by cut rather than on the average. For \gls{nmi}
and \gls{ari} it holds on all three log scales in both vocabularies at every evaluated cut but
seventeen, and those seventeen fall on $\log \psi_\gamma$ in the extended vocabulary. On
$\psi_\gamma$ it collapses, to $22$ and $35$ of $188$. On these two metrics the effect is therefore
carried by the representation rather than by the read-out scale: it survives every transformation
that leaves the distribution with a usable Euclidean metric, and fails only on the one that does
not. Mean per-cluster purity is less well behaved, and is taken up below.

\begin{table}[htbp]
\centering
\small
\caption{Number of evaluated cluster counts at which the ordering $\gamma = 0.2 > 0.5 > 0.8$ holds,
by feature scale, metric and vocabulary. Cuts are evaluated from $K = 3$ on the NVA subset ($98$
resolutions) and from $K = 13$ on the extended vocabulary ($188$).}
\label{tab:scale_horizon}
\begin{tabular}{lcccccc}
\toprule
& \multicolumn{3}{c}{NVA (of $98$)} & \multicolumn{3}{c}{Extended (of $188$)} \\
\cmidrule(lr){2-4}\cmidrule(lr){5-7}
Scale & Purity & \gls{nmi} & \gls{ari} & Purity & \gls{nmi} & \gls{ari} \\
\midrule
$z$ (reported)              & $98$ & $98$ & $98$ & $118$ & $188$ & $188$ \\
$\mathrm{clr}(\psi_\gamma)$ & $98$ & $98$ & $98$ & $160$ & $188$ & $188$ \\
$\log \psi_\gamma$          & $98$ & $98$ & $98$ & $185$ & $188$ & $171$ \\
$\psi_\gamma$               & $62$ & $89$ & $87$ & $96$  & $22$  & $35$  \\
\bottomrule
\end{tabular}
\end{table}

The two exceptions in the \gls{nmi} and \gls{ari} columns have different characters. The seventeen
failures of $\log \psi_\gamma$ on the extended vocabulary lie in one contiguous band, $K = 145$ to
$165$, where the two heads are separated by a median of $0.0004$ in \gls{ari}: these are ties rather
than inversions. Those of $\psi_\gamma$ are systematic: averaged over the scan
the $\gamma = 0.5$ head outscores $\gamma = 0.2$ there, at $0.2184$ against $0.2029$ in \gls{ari} and
$0.4427$ against $0.4334$ in \gls{nmi}. The reason is visible in Table~\ref{tab:scale_variance}. On
the raw scale the concentration of variance is itself a function of the horizon, the ten most
variable entries carrying $0.661$ of the total at $\gamma = 0.2$ but $0.612$ at $\gamma = 0.8$. The shortest horizon
produces the most sharply peaked distributions and is therefore the most distorted by a Euclidean
metric on that scale, so lengthening the horizon relieves the distortion at the same time as it
removes category information. Which of the two wins depends on how steep the underlying decline is.
On the NVA subset it is steep---on the reported read-out \gls{ari} falls by $0.060$ from
$\gamma = 0.2$ to $\gamma = 0.5$---and the ordering survives on $\psi_\gamma$ at all but the coarsest
resolutions, its failures confined to $K \le 43$. On the extended vocabulary it is about half as
steep, at $0.031$, and the relief outweighs it at almost every cut. None of this bears on the horizon
effect itself, which is measured on the log scale throughout.

Mean per-cluster purity behaves differently, and its failures belong to the extended vocabulary
rather than to any feature scale: it inverts at $70$ of the $188$ cuts even on the reported read-out,
and $69$ of those $70$ are inversions between the two closest heads. The cause is an interaction
between that metric and the composition of the extended vocabulary. Mean per-cluster purity weights
every cluster equally regardless of size. Seven of the thirteen categories there hold fewer than
sixty tokens, and these closed-class items keep distinctive successor profiles as the horizon
lengthens: from $\gamma = 0.2$ to $\gamma = 0.8$ the share of a category's tokens lying in its modal
cluster falls by $0.092$ on average for those seven, against $0.156$ for
the six categories holding two hundred; for determiners and pronouns it does not fall at all. High resolutions then fragment
the vocabulary---at $K = 200$, $119$ of the $200$ clusters hold three tokens or fewer, and such a
cluster is pure almost by construction. Both effects load the metric onto the part of the vocabulary
least sensitive to the horizon, leaving the $\gamma = 0.2$ and $\gamma = 0.5$ heads separated by a
mean of $0.0066$ across the scan, close enough for the partition to decide the comparison. Excluding
clusters of three tokens or fewer, the mean separation rises to $0.0326$ and the ordering holds at
$179$ of the $188$ cuts; weighting each cluster by its size removes them altogether on the reported
read-out, at $188$ of $188$ (Appendix~\ref{app:unk_gamma}). The NVA subset shows none of this: its three categories are
equally sized, its clusters at the reported resolutions are large, and the two heads are separated
there by $0.071$ on average, at $K = 3$ by $0.88$ against $0.59$---far beyond what the purity of
small fragments can offset.

\paragraph{What the per-word constant carries.}
The remaining difference between $z$ and $\mathrm{clr}(\psi_\gamma)$ is the log-normaliser $c$, and it
deserves naming because the training objective does not determine it: the gradient of the \gls{kl}
loss with respect to a head's outputs is $\psi_\gamma - G_\gamma$, which sums to zero over the
vocabulary, so adding a constant to a row of $z$ leaves the loss unchanged. Its value is fixed by the
parameterisation, not by the target. It is nonetheless not noise. Its standard deviation across the
evaluated tokens falls with the horizon, from $0.40$ nats at $\gamma = 0.2$ to $0.12$ at
$\gamma = 0.8$, and it correlates at $-0.68$ with the entropy of $\psi_\gamma$: $c$ measures how
peaked a word's successor distribution is. Ordered by it, the tag set runs from conjunctions and
determiners, which admit many continuations, to numerals, auxiliaries and proper nouns, which admit
few. Retaining it helps at the short horizon and costs a little at the long one, where the
distributions have flattened: on the NVA subset at $K=3$, $z$ leads $\log \psi_\gamma$ by
$0.106$ at $\gamma=0.2$ and by $0.088$ at $\gamma=0.5$, and trails it by $0.045$ at $\gamma=0.8$.
Since $\mathrm{clr}(\psi_\gamma)$ discards $c$ entirely and reproduces the reported results, no claim
in this paper rests on it.

\section{Count-Based Baselines}
\label{app:baselines}

The Limitations of the main text note that no comparison model was trained on the same corpus and
vocabulary, so the specificity of the reported geometry to the \gls{sr} objective is not established.
This appendix supplies the half of that comparison that requires no training: vectors derived
directly from corpus counts, pushed through the same analysis pipeline and scored against the same
reference labels.

\subsection{Construction}

All baselines are counted on the token stream the network was trained on: the same preprocessed
WikiText-103 training split, the same 20{,}000-token vocabulary, and the same per-line truncation at
80 tokens, so no count crosses a line or article boundary. Three families are compared:

\begin{itemize}
    \item \textbf{Bigram.} The row-normalised conditional $P(w' \mid w)$ over adjacent tokens. This
      is the $\gamma \to 0$ limit of the \gls{sr}: the immediate successor and nothing beyond it.
    \item \textbf{\Gls{ppmi}.} Positive pointwise mutual information over a symmetric $\pm 5$ window,
      the conventional count-based distributional vector. It is symmetric, and therefore records
      which words share contexts but neither the direction nor the distance of the relation.
    \item \textbf{Analytic \gls{sr}.} The closed-form $M_\gamma = \sum_{k \ge 0} \gamma^k P^{k+1}$,
      evaluated by Neumann series from the same empirical transition matrix $P$ (residuals below
      $10^{-9}$ at every $\gamma$). This is the object the network is trained to approximate,
      computed exactly and without learning.
\end{itemize}

Each matrix is restricted to the evaluated token rows, reduced by \gls{pca} retaining 99.99\% of the
variance, and clustered by the same consensus procedure used throughout: 240 trials over the 24 prime
base resolutions from 3 to 97, average-linkage agglomerative aggregation of the co-association
matrix, cut at every $K$ in the scan range. Only the $K$-Means ensemble is run; the baselines are a
control on the features rather than on the clustering algorithm. Each baseline, and for comparison the
trained $\gamma = 0.2$ head, is clustered under four consensus seeds: $42$, as throughout, and the $0$,
$7$ and $123$ of Appendix~\ref{app:unk_seed}. All values in this appendix are means over the four
seeds; seed $42$ of the trained head is the reported solution.

Baselines are compared on a logarithmic scale, matching the read-out actually clustered
(Appendix~\ref{app:feature_scale}). Bigram and analytic \gls{sr} vectors are taken as $\log(x + 10^{-9})$,
since transitions never observed have probability zero; \gls{ppmi} is logarithmic by construction.
The scale is not incidental: on the raw probability scale a
few high-frequency successors carry most of the variance, and the same analytic \gls{sr} at
$\gamma = 0.2$ scores an \gls{ari} of $0.017$ untransformed against $0.639$ under a logarithm --- the
same information, differently distributed. The effect weakens as the horizon lengthens and the
distributions flatten, with the raw analytic \gls{sr} at $\gamma = 0.8$ reaching $0.423$.

\subsection{Results}

Table~\ref{tab:baselines} compares the baselines with the trained \gls{sr} head at the four showcase resolutions,
all at $\gamma = 0.2$, the horizon analysed in the main text.

\begin{widetable}[htbp]
\centering
\caption{\gls{ari} and mean cluster purity for count-based baselines and the trained \gls{sr}
head, $K$-Means consensus, at the four showcase resolutions. Values are means over four consensus
seeds; the smaller line gives the range of each value across them. Baselines are shown on the
logarithmic scale (see text).}
\label{tab:baselines}
\small
\providecommand{\rng}[2]{{\scriptsize $#1$--$#2$}}
\setlength{\tabcolsep}{5pt}
\begin{fitwidth}
\begin{tabular}{lcccccccc}
\toprule
 & \multicolumn{2}{c}{NVA $K=3$} & \multicolumn{2}{c}{NVA $K=30$}
 & \multicolumn{2}{c}{Ext.\ $K=13$} & \multicolumn{2}{c}{Ext.\ $K=120$} \\
\cmidrule(lr){2-3}\cmidrule(lr){4-5}\cmidrule(lr){6-7}\cmidrule(lr){8-9}
Features & \gls{ari} & Purity & \gls{ari} & Purity & \gls{ari} & Purity & \gls{ari} & Purity \\
\midrule
\gls{sr}, $\gamma = 0.2$ (trained)  & $0.670$ & $0.882$ & $0.149$ & $0.864$ & $0.412$ & $0.628$ & $0.138$ & $0.811$ \\
                                    & \rng{0.659}{0.676} & \rng{0.880}{0.885} & \rng{0.145}{0.160} & \rng{0.849}{0.878} & \rng{0.397}{0.424} & \rng{0.615}{0.639} & \rng{0.133}{0.143} & \rng{0.806}{0.821} \\
\midrule
Analytic \gls{sr}, $\gamma = 0.2$   & $0.639$ & $0.869$ & $0.347$ & $0.911$ & $0.323$ & $0.662$ & $0.213$ & $0.809$ \\
                                    & \rng{0.571}{0.666} & \rng{0.850}{0.877} & \rng{0.342}{0.351} & \rng{0.901}{0.916} & \rng{0.317}{0.328} & \rng{0.631}{0.685} & \rng{0.198}{0.225} & \rng{0.792}{0.816} \\
Bigram                              & $0.459$ & $0.829$ & $0.419$ & $0.980$ & $0.179$ & $0.631$ & $0.306$ & $0.903$ \\
                                    & \rng{0.293}{0.519} & \rng{0.788}{0.845} & \rng{0.380}{0.442} & \rng{0.964}{0.985} & \rng{0.176}{0.182} & \rng{0.613}{0.670} & \rng{0.303}{0.308} & \rng{0.894}{0.909} \\
\Gls{ppmi}                          & $0.008$ & $0.494$ & $0.023$ & $0.708$ & $0.124$ & $0.452$ & $0.090$ & $0.693$ \\
                                    & \rng{0.001}{0.011} & \rng{0.482}{0.520} & \rng{0.019}{0.028} & \rng{0.697}{0.726} & \rng{0.108}{0.146} & \rng{0.432}{0.495} & \rng{0.084}{0.094} & \rng{0.684}{0.702} \\
\bottomrule
\end{tabular}
\end{fitwidth}
\end{widetable}

\paragraph{At the resolution of the reference categories, the trained head leads and is stable.}
On the extended vocabulary at $K = 13$ it reaches an \gls{ari} of $0.412$, against $0.323$ for the
analytic \gls{sr} and $0.179$ for the bigram, and leads under every seed. On the NVA subset at
$K = 3$ it reaches $0.670$, against $0.639$ and $0.459$. The margin over the analytic \gls{sr} is
small here, but the trained head varies by less than $0.02$ across seeds, while the count-based
features switch between distinct partitions (analytic \gls{sr} $0.571$ to $0.666$, bigram $0.293$
to $0.519$). Count-based vectors from the same corpus and vocabulary therefore do not recover the
word classes as reliably. The \gls{ppmi} vectors fail at every resolution, as expected of a
symmetric window, which discards the direction of the transitions the \gls{sr} encodes.

\paragraph{At the finest resolutions the bigram agrees better with the tag set, and this is informative.}
At $K = 30$ on the NVA subset the bigram reaches an \gls{ari} of $0.419$ against $0.149$ for the
trained head, and at $K = 120$ on the extended vocabulary $0.306$ against $0.138$. All methods are
cut to the same $K$, so the number of clusters cannot explain the gap. Purity can. The bigram's
clusters are almost entirely of one \gls{pos} tag ($0.980$ at $K = 30$, $0.903$ at $K = 120$): it
subdivides \emph{within} the reference categories. The trained head's are not ($0.864$ and $0.811$;
its least pure cluster at $K = 30$ reaches $0.399$): it forms groupings that \emph{cross} them, and
\gls{ari} penalises any partition organised by something other than the tag set. The gap thus
measures the category-crossing structure described in the Discussion of the main text, which
otherwise rests on individual examples such as the cluster combining duration nouns with aspectual
verbs. It is distinct from the decline of \gls{ari} with $K$ discussed there, which follows one
solution across resolutions rather than comparing several at one. Purity does not track \gls{ari} in
general: at $K = 13$ the trained head leads on \gls{ari} while its purity is level with the bigram
and below the analytic \gls{sr}.

\paragraph{The analytic \gls{sr} shows what multi-step accumulation adds to a bigram.}
$M_\gamma$ is computed from the bigram matrix $P$ alone, so it contains no information that $P$
lacks; its powers only redistribute the mass. That redistribution helps but does not account for
everything: on the extended vocabulary at $K = 13$ the bigram reaches $0.179$, the analytic \gls{sr}
$0.323$ and the trained head $0.412$. The analytic \gls{sr} is the $\lambda = 0$ end of the
interpolation in Appendix~\ref{app:methods_lambda}, and the limit a one-step target would converge
to, as noted in the main text. The trained model uses $\lambda = 0.9$, near the other end, where the
target keeps dependencies that no first-order chain represents. Its remaining margin over the
analytic \gls{sr} therefore combines two effects that this comparison cannot separate: that
non-Markovian structure, and the smoothing imposed by a shared $512$-dimensional trunk serving all
$20{,}000$ tokens. Separating them would need the empirical $k$-step occupancy (co-occurrences
counted at each distance and weighted by $\gamma^k$), the $\lambda = 1$ analogue, which we leave to
future work.

At the other two horizons the analytic \gls{sr} declines less steeply than the trained heads: on the
extended vocabulary at $K = 13$ it scores $0.323$, $0.331$ and $0.251$ at $\gamma = 0.2$, $0.5$ and
$0.8$, against $0.412$, $0.302$ and $0.214$. We draw no conclusion from this. The two families differ
in $\lambda$ as well as in being learned, and the logarithm that makes the scales comparable at one
horizon need not do so at another: untransformed, the analytic vectors score $0.017$ at
$\gamma = 0.2$ and $0.423$ at $\gamma = 0.8$ on the NVA subset. The horizon effect in the main text
concerns the trained heads only.

\end{appendices}
 \fi

\end{document}